\documentclass[10pt,twocolumn,letterpaper]{article}

\usepackage[camera]{cvpr}      

\usepackage{graphicx}
\usepackage{amsmath}
\usepackage{amssymb}
\usepackage{booktabs}
\usepackage{adjustbox}
\usepackage{xcolor}
\usepackage{abstract}
\usepackage{multirow}
\newcommand*\samethanks[1][\value{footnote}]{\footnotemark[#1]}

\usepackage[pagebackref,breaklinks,colorlinks]{hyperref}

\usepackage[capitalize]{cleveref}
\crefname{section}{Sec.}{Secs.}
\Crefname{section}{Section}{Sections}
\Crefname{table}{Table}{Tables}
\crefname{table}{Tab.}{Tabs.}

\def\cvprPaperID{8836} 
\def\confName{CVPR}
\def\confYear{2022}

\begin{document}

\title{\textcolor[rgb]{0.965,0.325,0.078}{N}\textcolor[rgb]{0.486,0.733,0}{Ü}\textcolor[rgb]{0,0.631,0.945}{W}\textcolor[rgb]{1,0.733,0}{A}: Visual Synthesis Pre-training for \textcolor[rgb]{0.965,0.325,0.078}{N}eural vis\textcolor[rgb]{0.486,0.733,0}{U}al \textcolor[rgb]{0,0.631,0.945}{W}orld cre\textcolor[rgb]{1,0.733,0}{A}tion}

\author{Chenfei Wu$^{1}$\thanks{Both authors contributed equally to this research.} \quad Jian Liang$^{2}$\samethanks[1] \quad Lei Ji$^{1}$ \quad  Fan Yang$^{1}$ \quad Yuejian Fang$^{2}$ \quad Daxin Jiang$^{1}$ \quad Nan Duan$^{1}$\thanks{Corresponding author.} \\
 {\small $^{1}$Microsoft Research Asia \quad $^{2}$Peking University} \\
{\tt\small\{chewu,leiji,fanyang,djiang,nanduan\}@microsoft.com}~~ {\tt\small\{j.liang@stu,fangyj@ss\}.pku.edu.cn}}

\twocolumn[{
\renewcommand\twocolumn[1][]{#1}
\maketitle
\begin{center}
    \centering
    \captionsetup{type=figure}
    \includegraphics[width=0.85\textwidth]{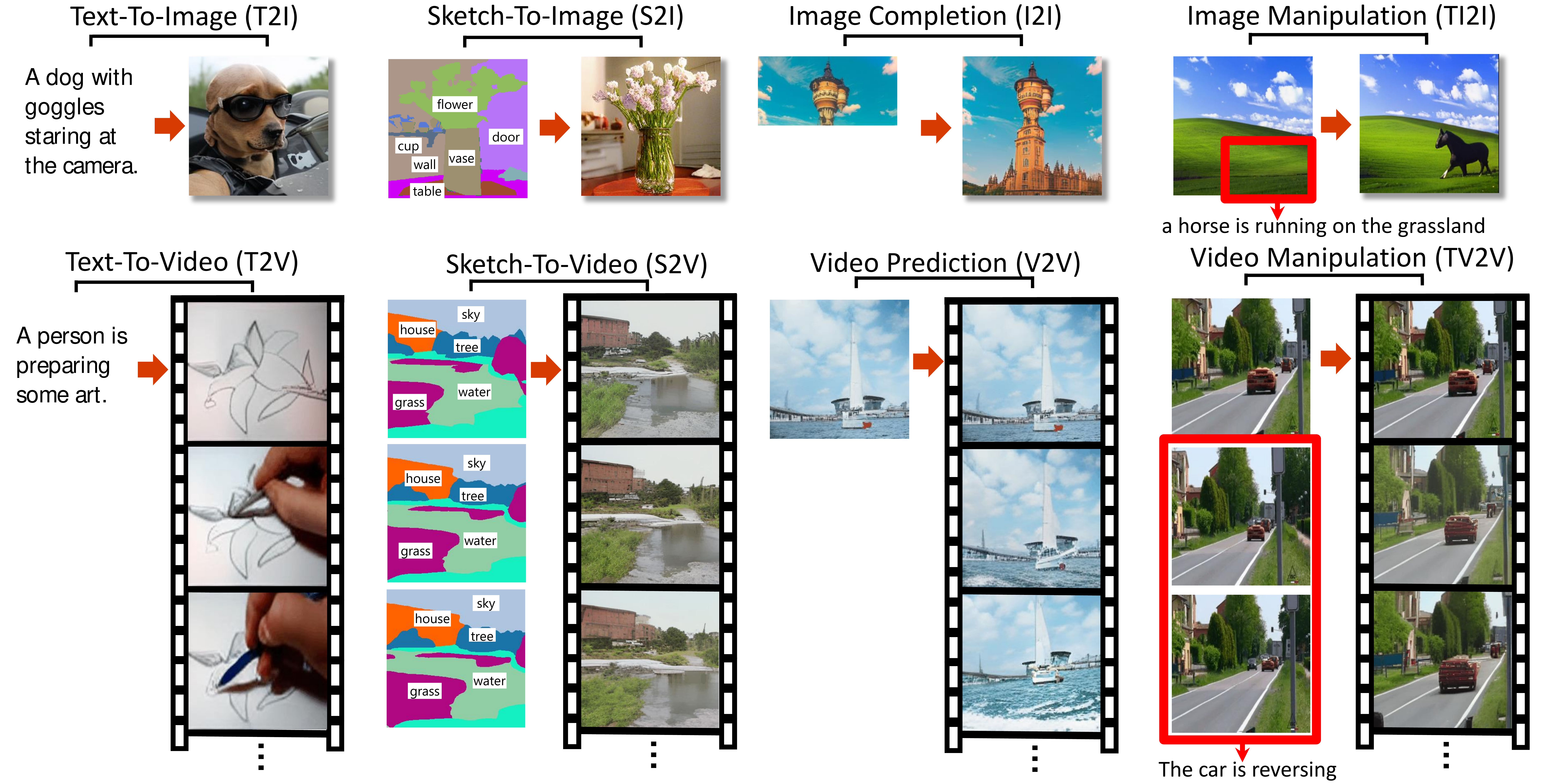}
    \captionof{figure}{Examples of 8 typical visual generation and manipulation tasks supported by the NÜWA model.}
    \label{fig:Overview}
    
\end{center}
}]
\saythanks
\begin{abstract}

This paper presents a unified multimodal pre-trained model called NÜWA that can generate new or manipulate existing visual data (i.e., images and videos) for various visual synthesis tasks. To cover language, image, and video at the same time for different scenarios, a 3D transformer encoder-decoder framework is designed, which can not only deal with videos as 3D data but also adapt to texts and images as 1D and 2D data, respectively. A 3D Nearby Attention (3DNA) mechanism is also proposed to consider the nature of the visual data and reduce the computational complexity. We evaluate NÜWA on 8 downstream tasks. Compared to several strong baselines, NÜWA achieves state-of-the-art results on text-to-image generation, text-to-video generation, video prediction, etc. Furthermore, it also shows surprisingly good zero-shot capabilities on text-guided image and video manipulation tasks. Project repo is \url{https://github.com/microsoft/NUWA}.

\end{abstract}

\section{Introduction}\label{sec:intro}

Nowadays, the Web is becoming more visual than ever before, as images and videos have become the new information carriers and have been used in many practical applications. With this background, visual synthesis is becoming a more and more popular research topic, which aims to build models that can generate new or manipulate existing visual data (i.e., images and videos) for various visual scenarios.

Auto-regressive models\cite{vanoordPixelRecurrentNeural2016,oordConditionalImageGeneration2016,wuGODIVAGeneratingOpenDomaIn2021,rameshZeroShotTexttoImageGeneration2021} play an important role in visual synthesis tasks, due to their explicit density modeling and stable training advantages compared with GANs\cite{brockLargeScaleGAN2019,tulyakovMocoganDecomposingMotion2018,xuAttnganFinegrainedText2018,radfordUnsupervisedRepresentationLearning2015}. 
Earlier visual auto-regressive models, such as PixelCNN\cite{oordConditionalImageGeneration2016}, PixelRNN\cite{vanoordPixelRecurrentNeural2016}, Image Transformer\cite{parmarImageTransformer2018}, iGPT\cite{chenGenerativepretrainingPixels2020}, and Video Transformer\cite{weissenbornScalingAutoregressiveVideo2020}, performed visual synthesis in a ``pixel-by-pixel'' manner.
However, due to their high computational cost on high-dimensional visual data, such methods can be applied to low-resolution images or videos only and are hard to scale up.

Recently, with the arise of VQ-VAE\cite{oordNeuralDiscreteRepresentation2017} as a discrete visual tokenization approach, efficient and large-scale pre-training can be applied to visual synthesis tasks for images (e.g., DALL-E\cite{rameshZeroShotTexttoImageGeneration2021} and CogView\cite{dingCogViewMasteringTexttoImage2021}) and videos (e.g., GODIVA\cite{wuGODIVAGeneratingOpenDomaIn2021}). Although achieving great success, such solutions still have limitations -- they treat images and videos separately and focus on generating either of them. This limits the models to benefit from both image and video data.

In this paper, we present NÜWA, a unified multimodal pre-trained model that aims to support visual synthesis tasks for both images and videos, and conduct experiments on 8 downstream visual synthesis, as shown in Fig.~\ref{fig:Overview}.
The main contributions of this work are three-fold:

\begin{itemize}
    \item We propose NÜWA, a general 3D transformer encoder-decoder framework, which covers language, image, and video at the same time for different visual synthesis tasks. It consists of an adaptive encoder that takes either text or visual sketch as input, and a decoder shared by 8 visual synthesis tasks.
    \item We propose a 3D Nearby Attention (3DNA) mechanism in the framework to consider the locality characteristic for both spatial and temporal axes. 3DNA not only reduces computational complexity but also improves the visual quality of the generated results.
    
    \item Compared to several strong baselines, NÜWA achieves state-of-the-art results on text-to-image generation, text-to-video generation, video prediction, etc. Furthermore, NÜWA shows surprisingly good zero-shot capabilities not only on text-guided image manipulation, but also text-guided video manipulation.
    
\end{itemize}

    

\section{Related Works}\label{sec:rw}

\subsection{Visual Auto-Regressive Models}

The method proposed in this paper follows the line of visual synthesis research based on auto-regressive models. Earlier visual auto-regressive models \cite{oordConditionalImageGeneration2016, vanoordPixelRecurrentNeural2016, parmarImageTransformer2018, chenGenerativepretrainingPixels2020, weissenbornScalingAutoregressiveVideo2020} performed visual synthesis in a ``pixel-by-pixel'' manner.
However, due to the high computational cost when modeling high-dimensional data, such methods can be applied to low-resolution images or videos only, and are hard to scale up.

\begin{figure*}[h]
	\centering
	\includegraphics[width=\textwidth]{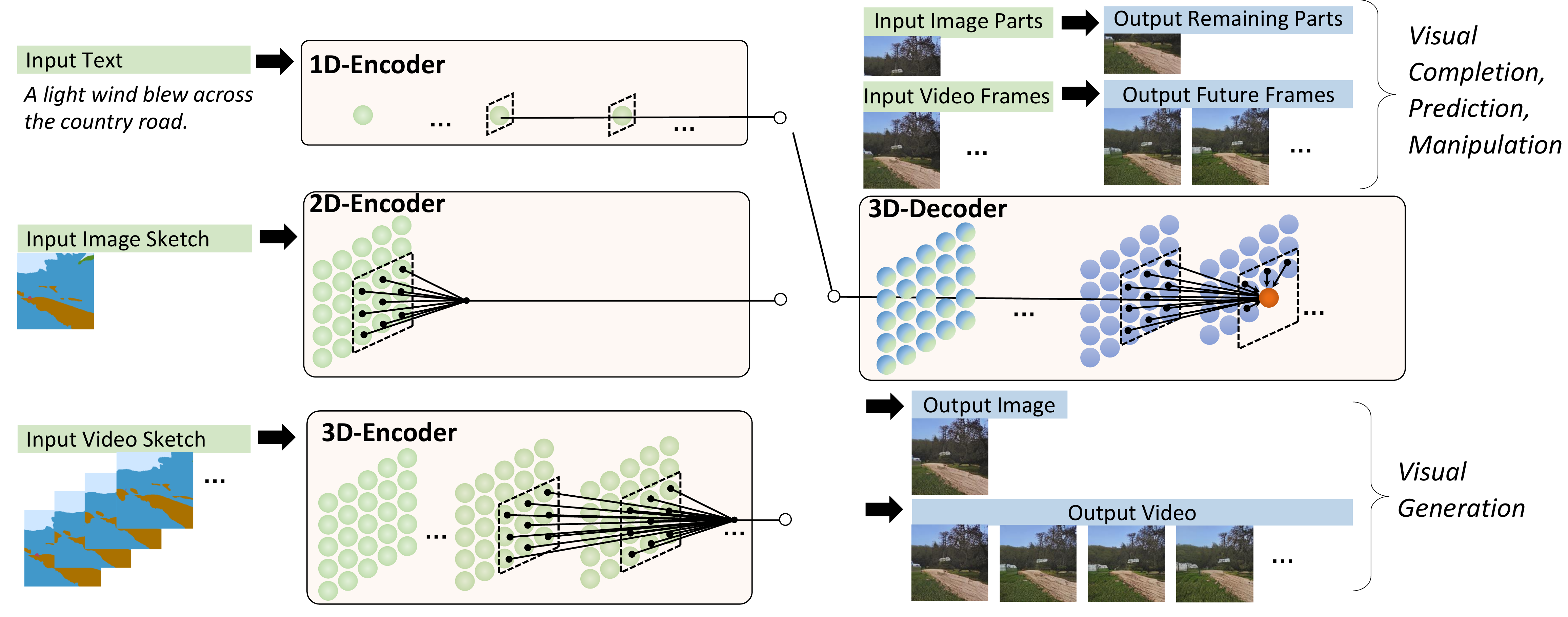}
	\caption{Overview structure of NÜWA. It contains an adaptive encoder supporting different conditions and a pre-trained decoder benefiting from both image and video data. For image completion, video prediction, image manipulation, and video manipulation tasks, the input partial images or videos are fed to the decoder directly.}
	\label{fig:Pipeline}
\end{figure*}
Recently, VQ-VAE-based~\cite{oordNeuralDiscreteRepresentation2017} visual auto-regressive models were proposed for visual synthesis tasks. By converting images into discrete visual tokens, such methods can conduct efficient and large-scale pre-training for text-to-image generation (e.g., DALL-E\cite{rameshZeroShotTexttoImageGeneration2021} and CogView\cite{dingCogViewMasteringTexttoImage2021}), text-to-video generation (e.g., GODIVA\cite{wuGODIVAGeneratingOpenDomaIn2021}), and video prediction (e.g., LVT\cite{rakhimovLatentVideoTransformer2020} and VideoGPT\cite{yanVideoGPTVideoGeneration2021}), with higher resolution of generated images or videos. 
However, none of these models was trained by images and videos together. But it is intuitive that these tasks can benefit from both types of visual data. 

Compared to these works, NÜWA is a unified auto-regressive visual synthesis model that is pre-trained by the visual data covering both images and videos and can support various downstream tasks. We also verify the effectiveness of different pretraining tasks in Sec.~\ref{sec:abl}. Besides, VQ-GAN\cite{esserTamingTransformersHighResolution2021} instead of VQ-VAE is used in NÜWA for visual tokenization, which, based on our experiment, can lead to better generation quality.

\subsection{Visual Sparse Self-Attention}

How to deal with the quadratic complexity issue brought by self-attention is another challenge, especially for tasks like high-resolution image synthesis or video synthesis.

Similar to NLP, sparse attention mechanisms have been explored to alleviate this issue for visual synthesis.
\cite{weissenbornScalingAutoregressiveVideo2020, rakhimovLatentVideoTransformer2020} split the visual data into different parts (or blocks) and then performed block-wise sparse attention for the synthesis tasks. 
However, such methods dealt with different blocks separately and did not model their relationships.
\cite{hoAxialAttentionMultidimensional2019, rameshZeroShotTexttoImageGeneration2021, wuGODIVAGeneratingOpenDomaIn2021} proposed to use axial-wise sparse attention in visual synthesis tasks, which conducts sparse attention along the axes of visual data representations. This mechanism makes training very efficient and is friendly to large-scale pre-trained models like DALL-E\cite{rameshZeroShotTexttoImageGeneration2021}, CogView\cite{dingCogViewMasteringTexttoImage2021}, and GODIVA\cite{wuGODIVAGeneratingOpenDomaIn2021}. However, the quality of generated visual contents could be harmed due to the limited contexts used in self-attention.
\cite{parmarImageTransformer2018, ramachandranStandaloneSelfattentionVision2019, childGeneratingLongSequences2019} proposed to use local-wise sparse attention in visual synthesis tasks, which allows the models to see more contexts. But these works were for images only.

Compared to these works, NÜWA proposes a 3D nearby attention that extends the local-wise sparse attention to cover both images to videos. We also verify that local-wise sparse attention is superior to axial-wise sparse attention for visual generation in Sec.~\ref{sec:abl}.


\section{Method}
\subsection{3D Data Representation} \label{sec:rep}
To cover all texts, images, and videos or their sketches, we view all of them as tokens and define a unified 3D notation $X\in \mathbb{R}^{h\times w
\times s\times d}$, where $h$ and $w$ denote the number of tokens in the spatial axis (height and width respectively), $s$ denotes the number of tokens in the temporal axis, and $d$ is the dimension of each token. In the following, we introduce how we get this unified representation for different modalities.

Texts are naturally discrete, and following Transformer\cite{vaswaniAttentionAllYou2017}, we use a lower-cased byte pair encoding (BPE) to tokenize and embed them into $\mathbb{R}^{1\times 1\times s\times d}$. We use placeholder 1 because the text has no spatial dimension.

Images are naturally continuous pixels. Input a raw image $I\in \mathbb{R}^{H\times W\times C}$ with height $H$, width $W$ and channel $C$,  VQ-VAE\cite{oordNeuralDiscreteRepresentation2017} trains a learnable codebook to build a bridge between raw continuous pixels and discrete tokens, as denoted in Eq.~(\ref{eq:zi})$\sim$(\ref{eq:I}):
\begin{equation} \label{eq:zi} 
z_i=\mathop{\arg\min}_{j}||E(I)_i-B_j||^2,
\end{equation}
\begin{equation} \label{eq:I} 
\hat{I}=G(B[z]),
\end{equation}
where $E$ is an encoder that encodes $I$ into $h\times w$ grid features $E(I)\in \mathbb{R}^{h\times w\times d_B}$, $B\in \mathbb{R}^{N\times d_B}$ is a learnable codebook with $N$ visual tokens, where each grid of $E(I)$ is searched to find the nearest token. The searched result $z\in \{0,1,…,N-1\}^{h\times w}$ are embedded by $B$ and reconstructed back to $\hat{I}$ by a decoder $G$. The training loss of VQ-VAE can be written as Eq.~(\ref{eq:lv}):
\begin{equation} \label{eq:lv} 
L^{V}=||I-\hat{I}||_2^2+||sg[E(I)]-B[z]||_2^2+||E(I)-sg[B[z]]||_2^2,
\end{equation}
where $||I-\hat{I}||_2^2$ strictly constraints the exact pixel match between $I$ and $\hat{I}$, which limits the generalization ability of the model. Recently, VQ-GAN\cite{esserTamingTransformersHighResolution2021} enhanced VQ-VAE training by adding a perceptual loss and a GAN loss to ease the exact constraints between $I$ and $\hat{I}$ and focus on high-level semantic matching, as denoted in Eq.~(\ref{eq:lp})$\sim$(\ref{eq:lg}):
\begin{equation} \label{eq:lp} 
L^{P}=||CNN(I)-CNN(\hat{I})||_2^2,
\end{equation}
\begin{equation} \label{eq:lg} 
L^{G}=logD(I)+log(1-D(\hat{I})).
\end{equation}
After the training of VQ-GAN, $B[z]\in \mathbb{R}^{h\times w\times 1\times d}$ is finally used as the representation of images. We use placeholder 1 since images have no temporal dimensions.

Videos can be viewed as a temporal extension of images, and recent works like VideoGPT\cite{yanVideoGPTVideoGeneration2021} and VideoGen\cite{zhangVideoGenGenerativeModeling2020} extend convolutions in the VQ-VAE encoder from 2D to 3D and train a video-specific representation. However, this fails to share a common codebook for both images and videos. In this paper, we show that simply using 2D VQ-GAN to encode each frame of a video can also generate temporal consistency videos and at the same time benefit from both image and video data. The resulting representation is denoted as $\mathbb{R}^{h\times w\times s
\times d}$, where $s$ denotes the number of frames.

For image sketches, we consider them as images with special channels. An image segmentation matrix $\mathbb{R}^{H\times W}$ with each value representing the class of a pixel can be viewed in a one-hot manner $\mathbb{R}^{H\times W\times C}$ where $C$ is the number of segmentation classes. By training an additional VQ-GAN for image sketch, we finally get the embedded image representation $\mathbb{R}^{h\times w\times 1\times d}$. Similarly, for video sketches, the representation is $R^{h\times w\times s\times d}$.

\subsection{3D Nearby Self-Attention}\label{sec:nearby}
In this section, we define a unified 3D Nearby Self-Attention (3DNA) module based on the previous 3D data representations, supporting both self-attention and cross-attention. We first give the definition of 3DNA in Eq.~(\ref{eq:Y}), and introduce detailed implementation in Eq.~(\ref{eq:nijk})$\sim$(\ref{eq:yijk}):
\begin{equation} \label{eq:Y} 
Y=3DNA(X, C;W),
\end{equation}
where both $X \in \mathbb{R}^{h\times w\times s\times d^{in}}$ and $C\in \mathbb{R}^{h'\times w'\times s'\times d^{in}}$ are 3D representations introduced in Sec.~\ref{sec:rep}. If $C=X$, 3DNA denotes the self-attention on target $X$ and if $C\neq X$ , 3DNA is cross-attention on target $X$ conditioned on $C$. $W$ denotes learnable weights.

We start to introduce 3DNA from a coordinate $(i,j,k)$ under $X$. By a linear projection, the corresponding coordinate $(i',j',k')$ under $C$ is $\left(\lfloor i\frac{h'}{h}\rfloor, \lfloor j\frac{w'}{w}\rfloor, \lfloor k\frac{s'}{s}\rfloor\right)$. Then, the local neighborhood around $(i',j',k')$ with a width, height and temporal extent $e^w,e^h,e^s\in \mathbb{R}^+$ is defined in Eq.~(\ref{eq:nijk}),
\begin{small}
\begin{equation} \label{eq:nijk}
N^{(i,j,k)}=\left\{C_{abc}\bigg| \left|a-i'\right|\leq e^h,\left|b-j'\right|\leq e^w,\left|c-k'\right|\leq e^s\right\},
\end{equation}
\end{small}
where $N^{(i,j,k)}\in \mathbb{R}^{e^h\times e^w\times e^s\times d^{in}}$ is a sub-tensor of condition $C$ and consists of the corresponding nearby information that $(i,j,k)$ needs to attend. With three learnable weights $W_Q,W_K,W_V\in \mathbb{R}^{d^{in}\times d^{out}}$, the output tensor for the position $(i,j,k)$ is denoted in Eq.~(\ref{eq:qijk})$\sim$(\ref{eq:yijk}):
\begin{gather}
Q^{(i,j,k)}=XW^Q \label{eq:qijk}  \\
K^{(i,j,k)}=N^{(i, j, k)}W^K \label{eq:kijk} \\
V^{(i,j,k)}=N^{(i, j, k)}W^V \label{eq:vijk}\\
y_{ijk}=softmax\left(\frac{(Q^{(i,j,k)})^\mathsf{T} K^{(i,j,k)}}{\sqrt{d^{in}}}\right)V^{(i,j,k)} \label{eq:yijk} 
\end{gather}  
where the $(i,j,k)$ position queries and collects corresponding nearby information in $C$. This also handles $C=X$, then $(i,j,k)$ just queries the nearby position of itself. 3NDA not only reduces the complexity of full attention from $O\left(\left(hws\right)^2\right)$ to $O\left(\left(hws\right)\left(e^he^we^s\right)\right)$,  but also shows superior performance and we discuss it in Sec.~\ref{sec:abl}.

\subsection{3D Encoder-Decoder} \label{sec:ed}
In this section, we introduce 3D encode-decoder built based on 3DNA. To generate a target $Y\in \mathbb{R}^{h\times w\times s\times d^{out}}$ under the condition of $C\in \mathbb{R}^{h'\times w'\times s'\times d^{in}}$, the positional encoding for both $Y$ and $C$ are updated by three different learnable vocabularies considering height, width, and temporal axis, respectively in Eq.~(\ref{eq:xijk})$\sim$(\ref{eq:cijk}): 
\begin{gather}
Y_{ijk}:=Y_{ijk}+P_i^h+P_j^w+P_k^s \label{eq:xijk} \\
C_{ijk}:=C_{ijk}+P_i^{h'}+P_j^{w'}+P_k^{s'} \label{eq:cijk}
\end{gather}  
Then, the condition $C$ is fed into an encoder with a stack of $L$ 3DNA layers to model the self-attention interactions, with the $l$th layer denoted in Eq.~(\ref{eq:encoder}):
\begin{equation} \label{eq:encoder} 
C^{(l)}=3DNA(C^{(l-1)},C^{(l-1)}),
\end{equation}
Similarly, the decoder is also a stack of $L$ 3DNA layers. The decoder calculates both self-attention of generated results and cross-attention between generated results and conditions. The $l$th layer is denoted in Eq.~(\ref{eq:decoder}).
\begin{equation} \label{eq:decoder}
\begin{aligned}
Y_{ijk}^{(l)}=&3DNA(Y_{<i,<j,<k}^{(l-1)}, Y_{<i,<j,<k}^{(l-1)})\\
+&3DNA(Y_{<i,<j,<k}^{(l-1)}, C^{(L)}),
\end{aligned}
\end{equation}
where $<i,<j,<k$ denote the generated tokens for now. The initial token $V_{0,0,0}^{(1)}$ is a special $<bos>$ token learned during the training phase.
\subsection{Training Objective}\label{sec:training}
We train our model on three tasks, Text-to-Image (T2I), Video Prediction (V2V) and Text-to-Video (T2V). The training objective for the three tasks are cross-entropys denoted as three parts in Eq.~(\ref{eq:loss}), respectively:
\begin{align} \label{eq:loss} 
\begin{split}
\mathcal{L}=&-\sum\nolimits_{t=1}^{h\times w}log~p_\theta\left(y_t\big|y_{<t},C^{text};\theta\right)\\
&-\sum\nolimits_{t=1}^{h\times w\times s}log~p_\theta\left(y_t\big|y_{<t},c;\theta\right)\\
&-\sum\nolimits_{t=1}^{h\times w\times s}log~p_\theta\left(y_t\big|y_{<t},C^{text};\theta\right)
\end{split}
\end{align}
For T2I and T2V tasks, $C^{text}$ denotes text conditions. For the V2V task, since there is no text input, we instead get a constant 3D representation $c$ of the special word ``None''. $\theta$ denotes the model parameters.

\section{Experiments} \label{sec:exp}
Based on Sec.~\ref{sec:training} we first pre-train NÜWA on three datasets: Conceptual Captions\cite{linMicrosoftCocoCommon2014} for text-to-image (T2I) generation, which includes 2.9M text-image pairs, 
Moments in Time\cite{monfortMomentsTimeDataset2019} for video prediction (V2V), which includes 727K videos, 
and VATEX dataset\cite{wangVatexLargescaleHighquality2019} for text-to-video (T2V) generation, which includes 241K text-video pairs. In the following, we first introduce implementation details in Sec.~\ref{sec:imp} and then compare NÜWA with state-of-the-art models in Sec.~\ref{sec:cmp}, and finally conduct ablation studies in Sec.~\ref{sec:abl} to study the impacts of different parts.



\subsection{Implementation Details} \label{sec:imp}
In Sec.~\ref{sec:rep}, we set the sizes of 3D representations for text, image, and video as follows.
For text, the size of 3D representation is $1\times 1\times 77\times 1280$. 
For image, the size of 3D representation is $21\times 21\times 1\times 1280$.
For video, the size of 3D representation is $21\times 21\times 10\times 1280$, where we sample 10 frames from a video with 2.5 fps. Although the default visual resolution is $336\times 336$, we pre-train different resolutions for a fair comparison with existing models.
For the VQ-GAN model used for both images and videos, the size of grid feature $E(I)$ in Eq.~(\ref{eq:zi}) is $441\times 256$, and the size of the codebook $B$ is $12,288$.

Different sparse extents are used for different modalities in Sec.~\ref{sec:nearby}. 
For text, we set $(e^w,e^h,e^s)=(1,1,\infty)$, where $\infty$ denotes that the full text is always used in attention. 
For image and image sketches, $(e^w,e^h,e^s)=(3,3,1)$. 
For video and video sketches, $(e^w,e^h,e^s)=(3,3,3)$. 

We pre-train on 64 A100 GPUs for two weeks with the layer $L$ in Eq.~(\ref{eq:encoder}) set to 24, an Adam~\cite{Kingma_Adammethodstochastic_2014} optimizer with a learning rate of 1e-3, a batch size of 128, and warm-up 5\% of a total of 50M steps. The final pre-trained model has a total number of 870M parameters.

\begin{figure*}[t]
	\centering
	\includegraphics[width=\textwidth]{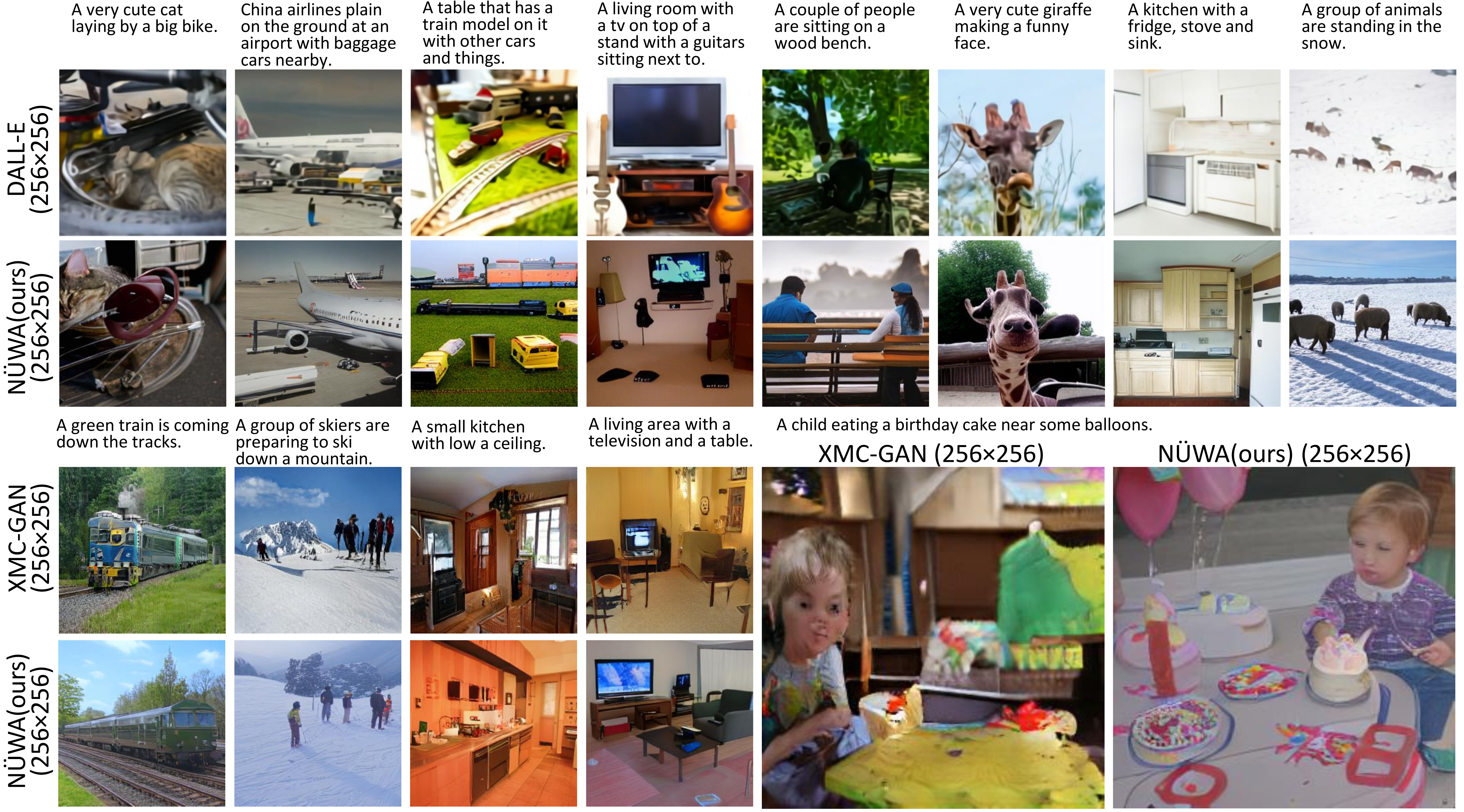}
	\caption{Qualitative comparison with state-of-the-art models for Text-to-Image (T2I) task on MSCOCO dataset.}
	\label{fig:T2I}
	\vspace{-2mm}
\end{figure*}
\begin{figure*}[t]
	\centering
	\includegraphics[width=\textwidth]{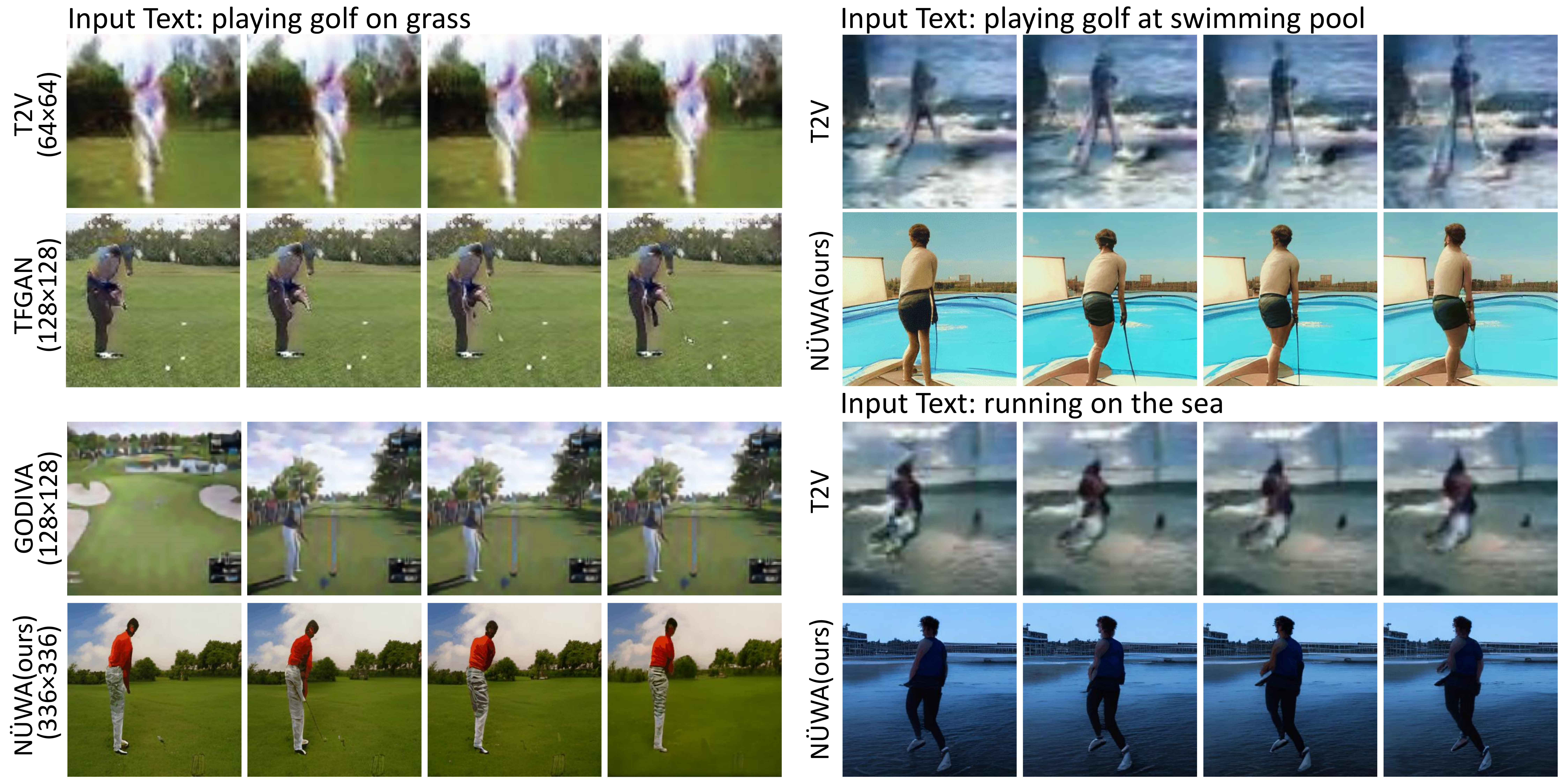}
	\caption{Quantitative comparison with state-of-the-art models for Text-to-Video (T2V) task on Kinetics dataset.}
	\label{fig:T2V_3}
	\vspace{-4mm}
\end{figure*}

\subsection{Comparison with state-of-the-art} \label{sec:cmp}
\vspace{-4mm}
\newcommand{\bwidth}{0.7cm}
\begin{table}[t]
\footnotesize
\caption{Qualitative comparison with the state-of-the-art models for Text-to-Image (T2I) task on the MSCOCO (256×256) dataset.}
\label{tab:t2i}
\tabcolsep=0.06cm
\begin{tabular}{p{1.9cm}p{0.9cm}p{\bwidth}p{\bwidth}p{\bwidth}p{0.7cm}p{\bwidth}p{1.1cm}}
\toprule
Model              & FID-0↓      & FID-1       & FID-2         & FID-4         & FID-8         & IS↑           & CLIPSIM↑ \\
\midrule
AttnGAN\cite{xuAttnganFinegrainedText2018}     & 35.2        & 44.0          & 72.0            & 108.0           & 100.0           & 23.3          & 0.2772        \\
DM-GAN\cite{zhuDmganDynamicMemory2019}     & 26.0          & 39.0          & 73.0            & 119.0           & 112.3         & \textbf{32.2} & 0.2838        \\
DF-GAN\cite{taoDfganDeepFusion2020}     & 26.0          & 33.8        & 55.9          & 91.0            & 97.0            & 18.7          & 0.2928        \\
  DALL-E\cite{rameshZeroShotTexttoImageGeneration2021}  & 27.5        & 28.0          & 45.5          & 83.5          & 85.0            & 17.9          & -        \\
CogView\cite{dingCogViewMasteringTexttoImage2021} & 27.1        & 19.4        & \textbf{13.9} & 19.4 & \textbf{23.6} & 18.2          & 0.3325        \\
XMC-GAN\cite{zhangCrossmodalContrastiveLearning2021} & \textbf{9.3}        & -        & - & -& - & 30.5          & -        \\

\midrule
NÜWA               & 12.9 & \textbf{13.8} & 15.7          & \textbf{19.3} & 24          & 27.2          & \textbf{0.3429}   \\
\bottomrule
\end{tabular}
\label{tab:metric}
\vspace{-3mm}
\end{table}

\newcommand{\swidth}{1cm}
\begin{table}[t]
\footnotesize
\begin{center}
\caption{Quantitative comparison with state-of-the-art models for Text-to-Video (T2V) task on Kinetics dataset.}
\label{tab:t2v}
\tabcolsep=0.11cm
\begin{tabular}{p{2.7cm}p{0.7cm}p{1.2cm}p{1.2cm}p{1.3cm}}
\toprule
Model             & Acc↑     & FID-img↓      & FID-vid↓    & CLIPSIM↑ \\
\midrule
T2V  (64×64) \cite{liVideoGenerationText2018}       & 42.6     & 82.13     & 14.65       & 0.2853        \\
SC  (128×128) \cite{balajiConditionalGANDiscriminative2019}        & 74.7     & 33.51     & 7.34        & 0.2915        \\
TFGAN (128×128)\cite{balajiConditionalGANDiscriminative2019}     & 76.2     & 31.76     & 7.19        & 0.2961        \\
\midrule
NÜWA (128×128)     & \textbf{77.9}     & \textbf{28.46}     & \textbf{7.05}         & \textbf{0.3012}        \\
\bottomrule
\end{tabular}
\end{center}
\vspace{-6mm}
\end{table}
\begin{table}[t]
\footnotesize
\begin{center}
\caption{Quantitative comparison with state-of-the-art models for Video Prediction (V2V) task on BAIR (64×64) dataset.}
\label{tab:v2v}
\begin{tabular}{p{4.6cm}p{1.3cm}p{1.3cm}}
\toprule
Model                            & Cond. & FVD↓  \\
\midrule
MoCoGAN\cite{tulyakovMocoganDecomposingMotion2018}                 & 4     & 503   \\
SVG-FP\cite{dentonStochasticVideoGeneration2018}                   & 2     & 315   \\
CNDA\cite{finnUnsupervisedLearningPhysical2016a}                     & 2     & 297   \\
SV2P\cite{babaeizadehStochasticVariationalVideo2017a}                     & 2     & 263   \\
SRVP\cite{franceschiStochasticLatentResidual2020}                  & 2     & 181   \\
VideoFlow\cite{kumarVideoflowConditionalFlowbased2019}                & 3     & 131   \\
LVT\cite{rakhimovLatentVideoTransformer2020}                      & 1     & 126±3 \\
SAVP\cite{leeStochasticAdversarialVideo2018} &2 & 116 \\
DVD-GAN-FP\cite{clarkWhatDoesBert2019}   & 1     & 110   \\
Video   Transformer (S)\cite{weissenbornScalingAutoregressiveVideo2020}  & 1     & 106±3 \\
TriVD-GAN-FP\cite{lucTransformationbasedAdversarialVideo2020}             & 1     & 103   \\
CCVS\cite{moingCCVSContextawareControllable2021}                  & 1     & 99±2  \\
Video   Transformer (L)\cite{weissenbornScalingAutoregressiveVideo2020}  & 1     & 94±2  \\
\midrule
NÜWA                             & 1     & \textbf{86.9}  \\
\bottomrule
\end{tabular}
\end{center}
\vspace{-5mm}
\end{table}

\begin{figure}[t]
	\centering
	\includegraphics[width=3.4in]{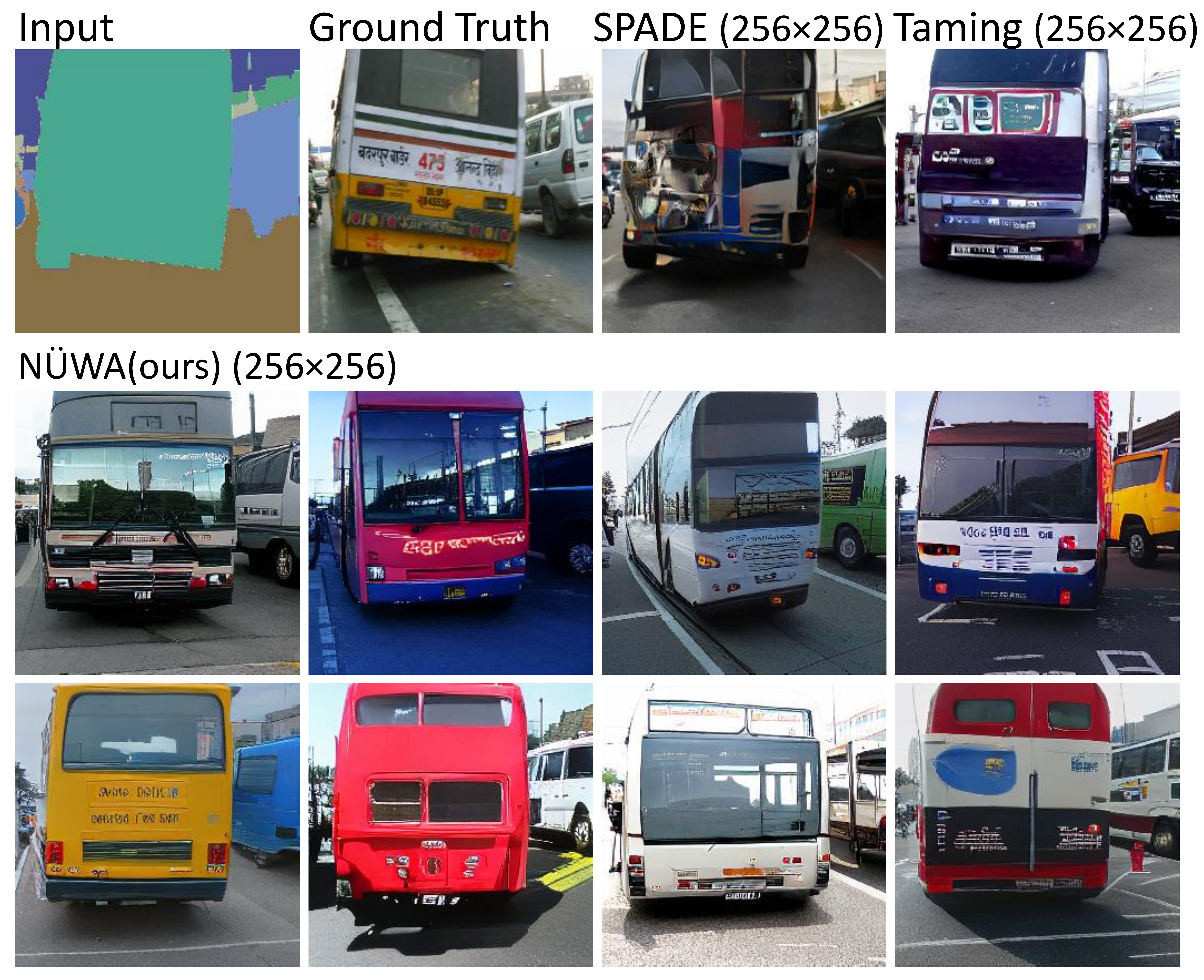}
	\caption{Quantitative comparison with state-of-the-art models for Sketch-to-Image (S2I) task on MSCOCO stuff dataset.}
	\label{fig:S2I}
	\vspace{-4mm}
\end{figure}
\begin{figure}[t]
	\centering
	\includegraphics[width=3.4in]{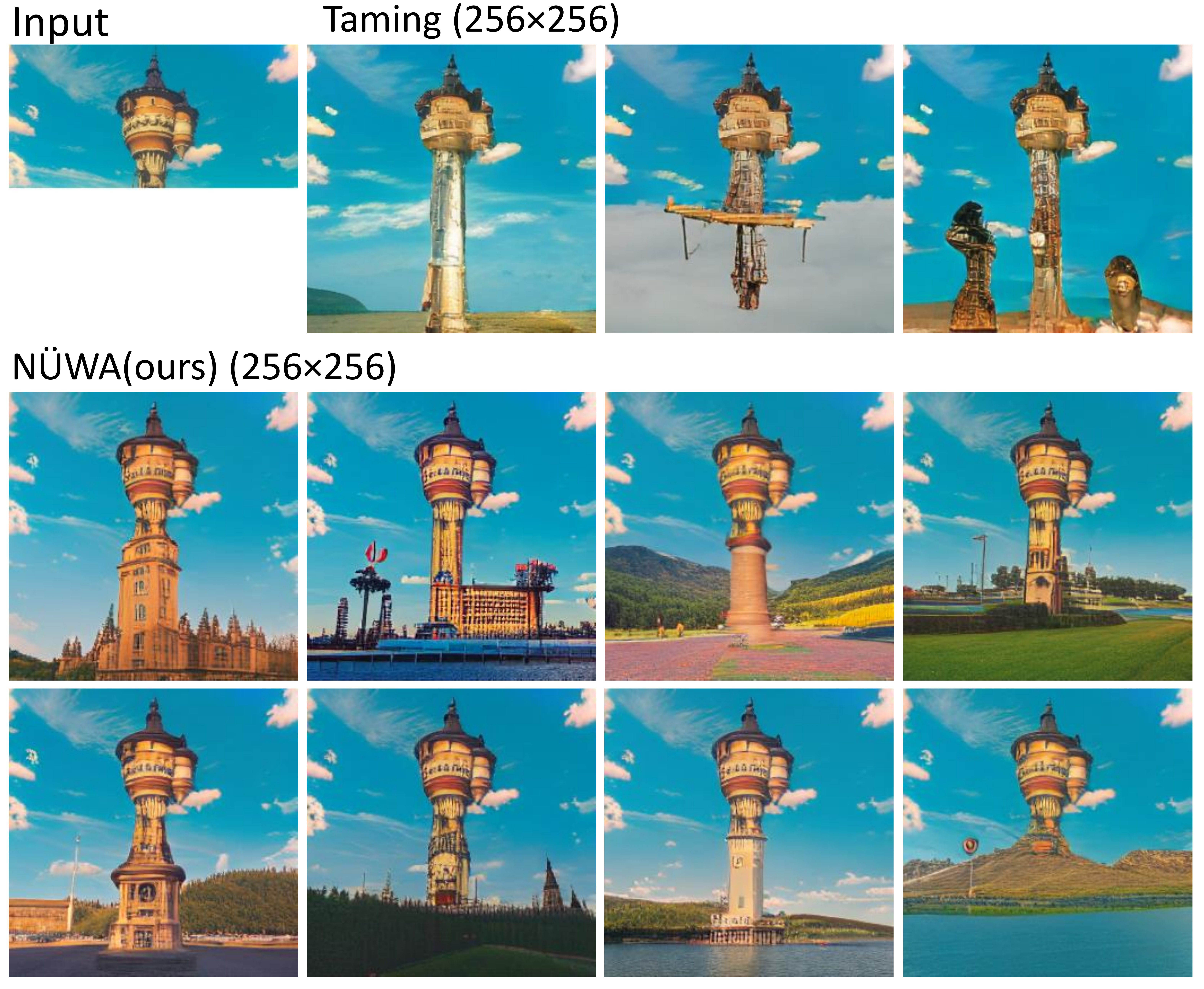}
	\caption{Qualitative comparison with the state-of-the-art model for Image Completion (I2I) task in a zero-shot manner.}
	\label{fig:I2I}
	\vspace{-5mm}
\end{figure}

\textbf{Text-to-Image (T2I) fine-tuning}: We compare NÜWA on the MSCOCO\cite{linMicrosoftCocoCommon2014} dataset quantitatively in Tab.~\ref{tab:t2i} and qualitatively in Fig.~\ref{fig:T2I}. Following DALL-E\cite{rameshZeroShotTexttoImageGeneration2021}, we use $k$ blurred FID score (FID-$k$) and Inception Score (IS)\cite{salimansImprovedTechniquesTraining2016} to evaluate the quality and variety respectively, and following GODIVA\cite{wuGODIVAGeneratingOpenDomaIn2021}, we use CLIPSIM metric, which incorporates a CLIP\cite{radfordLearningTransferableVisual2021} model to calculate the semantic similarity between input text and the generated image. For a fair comparison, all the models use the resolution of $256\times 256$. We generate 60 images for each text and select the best one by CLIP\cite{radfordLearningTransferableVisual2021}. In Tab.~\ref{tab:t2i}, NÜWA significantly outperforms CogView\cite{dingCogViewMasteringTexttoImage2021} with FID-0 of 12.9 and CLIPSIM of 0.3429. Although XMC-GAN\cite{zhangCrossmodalContrastiveLearning2021} reports a significant FID score of 9.3, we find NÜWA generates more realistic images compared with the exact same samples in XMC-GAN's paper (see Fig.~\ref{fig:T2I}). Especially in the last example, the boy's face is clear and the balloons are correctly generated. 



\textbf{Text-to-Video (T2V) fine-tuning}: We compare NÜWA on the Kinetics\cite{kayKineticsHumanAction2017} dataset quantitatively in Tab.~\ref{tab:t2v} and qualitatively in Fig.~\ref{fig:T2V_3}. Following TFGAN\cite{balajiConditionalGANDiscriminative2019}, we evaluate the visual quality on FID-img and FID-vid metrics and semantic consistency on the accuracy of the label of generated video. As shown in Tab.~\ref{tab:t2v}, NÜWA achieves the best performance on all the above metrics. In Fig.~\ref{fig:T2V_3}, we also show the strong zero-shot ability for generating unseen text, such as “playing golf at swimming pool” or “running on the sea”.

\textbf{Video Prediction (V2V) fine-tuning}: We compare NÜWA on BAIR Robot Pushing\cite{ebertSelfSupervisedVisualPlanning2017} dataset quantitatively in Tab.~\ref{tab:v2v}. Cond. denotes the number of frames given to predict future frames. For a fair comparison, all the models use 64×64 resolutions. Although given only one frame as condition (Cond.), NÜWA still significantly pushes the state-of-the-art FVD\cite{unterthinerAccurateGenerativeModels2018} score from 94±2 to 86.9.

\textbf{Sketch-to-Image (S2I) fine-tuning}: We compare NÜWA on MSCOCO stuff\cite{linMicrosoftCocoCommon2014} qualitatively in Fig.~\ref{fig:S2I}. NÜWA generates realistic buses of great varieties compared with Taming-Transformers\cite{esserTamingTransformersHighResolution2021} and SPADE\cite{parkSemanticImageSynthesis2019}. Even the reflection of the bus window is clearly visible.

\begin{figure}[t]
	\centering
	\includegraphics[width=3.4in]{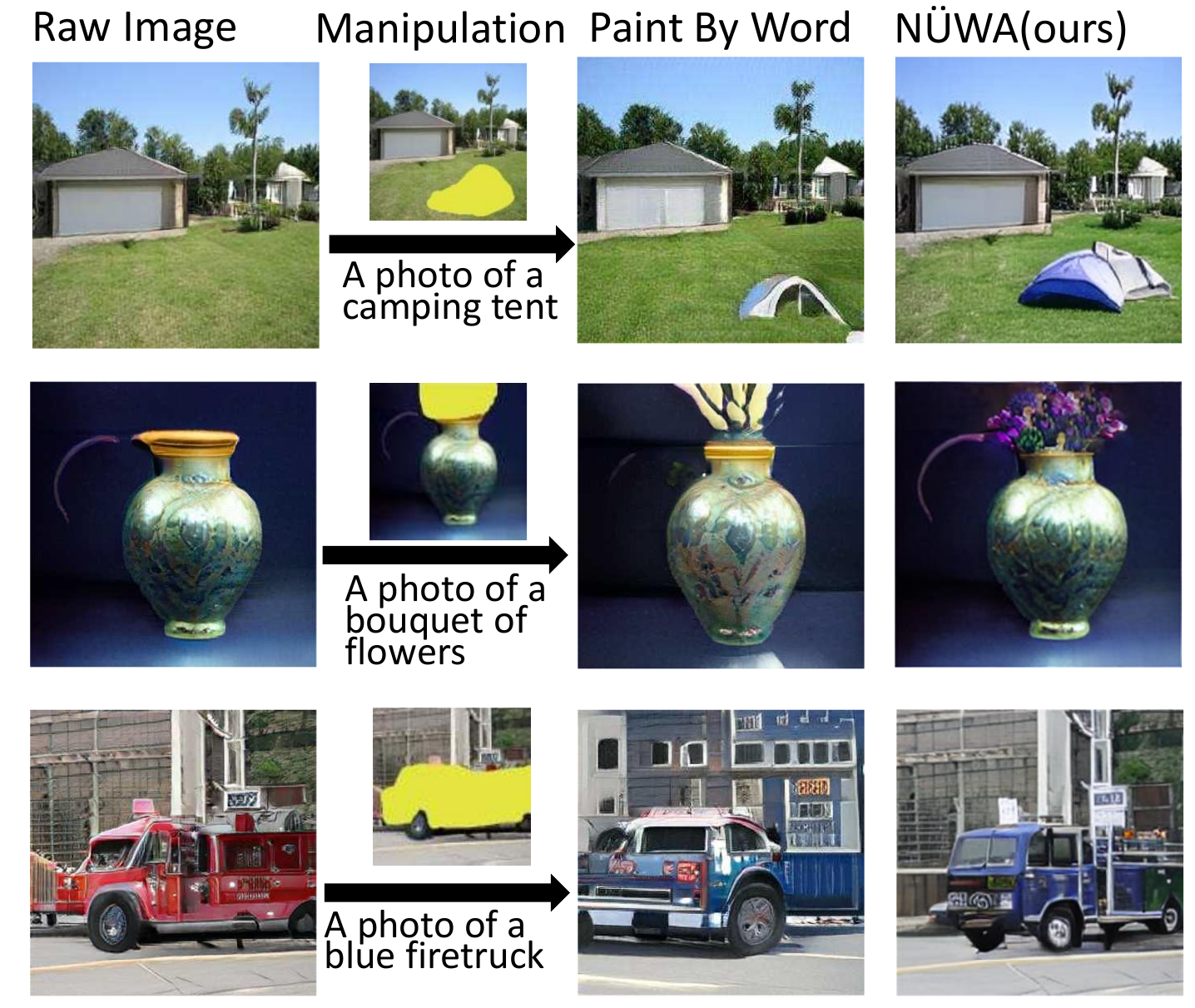}
	\caption{Quantitative comparison with state-of-the-art models for text-guided image manipulation (TI2I) in a zero-shot manner.}
	\label{fig:TI2I}
	\vspace{-2mm}
\end{figure}

\textbf{Image Completion (I2I) zero-shot evaluation}: We compare NÜWA in a zero-shot manner qualitatively in Fig.~\ref{fig:I2I}. Given the top half of the tower, compared with Taming Transformers\cite{esserTamingTransformersHighResolution2021}, NÜWA shows richer imagination of what could be for the lower half of the tower, including buildings, lakes, flowers, grass, trees, mountains, etc.

\textbf{Text-Guided Image Manipulation (TI2I) zero-shot evaluation}: We compare NÜWA in a zero-shot manner qualitatively in Fig.~\ref{fig:TI2I}. Compared with Paint By Word \cite{bauPaintWord2021}, NÜWA shows strong manipulation ability, generating high-quality text-consistent results while not changing other parts of the image. For example, in the third row, the blue firetruck generated by NÜWA is more realistic, while the behind buildings show no change. This is benefited from real-world visual patterns learned by multi-task pre-training on various visual tasks. Another advantage is the inference speed of NÜWA, practically 50 seconds to generate an image, while Paint By Words requires additional training during inference, and takes about 300 seconds to converge.

\textbf{Sketch-to-Video (S2V) fine-tuning} and \textbf{Text-Guided Video Manipulation (TV2V) zero-shot evaluation}: As far as we know, open-domain S2V and TV2V are tasks first proposed in this paper. Since there is no comparison, we instead arrange them in Ablation Study in Section~\ref{sec:abl}.

More detailed comparisons, samples, including human evaluations, are provided in the appendix.

\subsection{Ablation Study}\label{sec:abl}


The above part of Tab.~\ref{tab:VQ-GAN} shows the effectiveness of different VQ-VAE (VQ-GAN) settings. We experiment on ImageNet\cite{russakovskyImageNetLargeScale2015} and OpenImages\cite{kuznetsovaOpenImagesDataset2020}. $R$ denotes raw resolution, $D$ denotes the number of discrete tokens. The compression rate is denoted as $Fx$, where $x$ is the quotient of $\sqrt{R}$ divided by $\sqrt{D}$. Comparing the first two rows in Tab.~\ref{tab:VQ-GAN}, VQ-GAN shows significantly better Fréchet Inception Distance (FID)\cite{heuselGansTrainedTwo2017} and Structural Similarity Matrix (SSIM) scores than VQ-VAE. Comparing Row 2-3, we find that the number of discrete tokens is the key factor leading to higher visual quality instead of compress rate. Although Row 2 and Row 4 have the same compression rate F16, they have different FID scores of 6.04 and 4.79. So what matters is not only how much we compress the original image, but also how many discrete tokens are used for representing an image. This is in line with cognitive logic, it’s too ambiguous to represent human faces with just one token. And practically, we find that $16^2$ discrete tokens usually lead to poor performance, especially for human faces, and $32^2$ tokens show the best performance. However, more discrete tokens mean more computing, especially for videos. We finally use a trade-off version for our pre-training: $21^2$ tokens. By training on the Open Images dataset, we further improve the FID score of the $21^2$ version from 4.79 to 4.31.


\begin{figure}[t]
	\centering
	\includegraphics[width=3.4in]{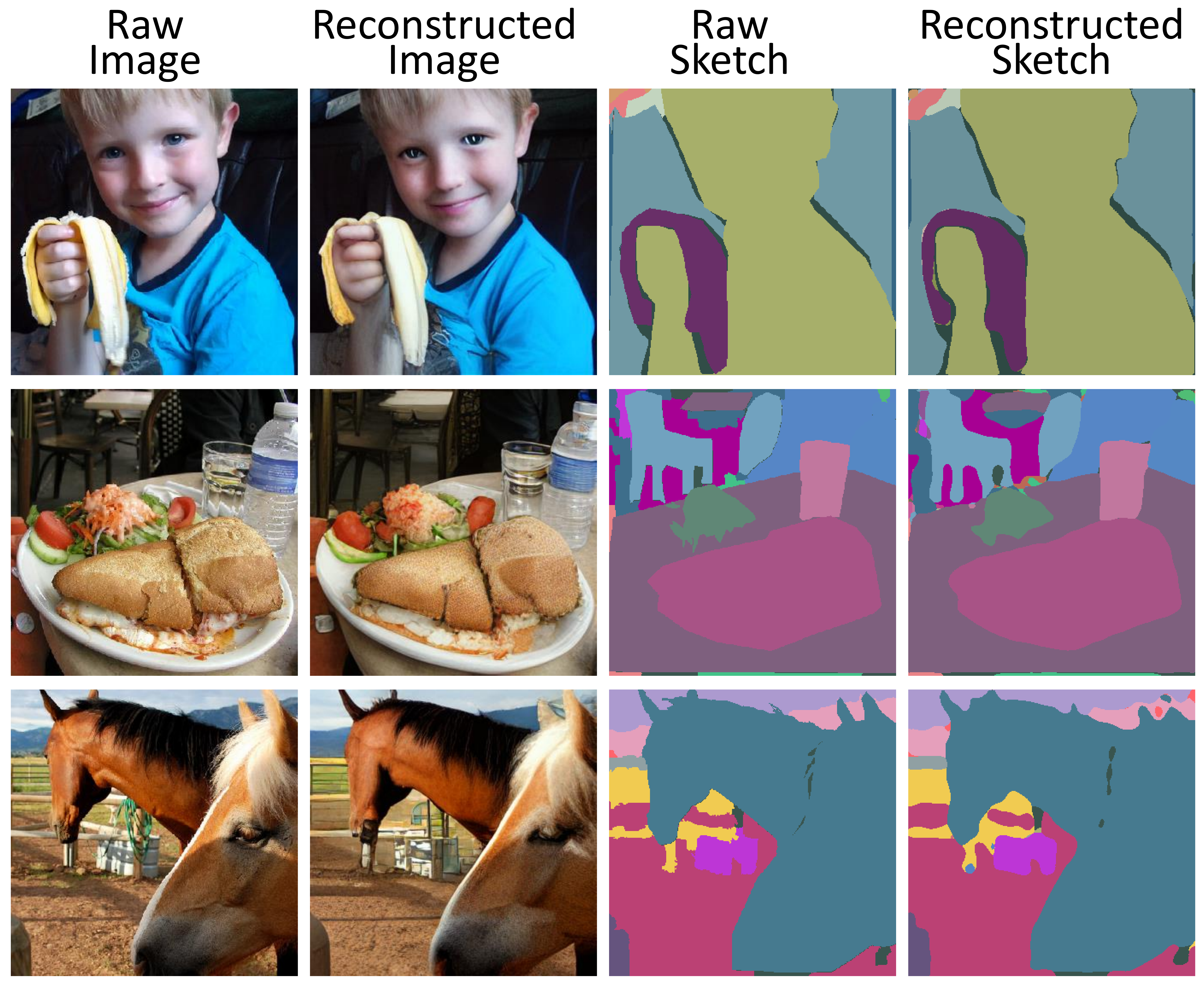}
	\caption{Reconstruction samples of VQ-GAN and VQ-GAN-Seg.}
	\label{fig:VQ-GANSketch}
	\vspace{-2mm}
\end{figure}

\begin{table}[t]
\footnotesize
\begin{center}
\caption{Effectiveness of different VQ-VAE (VQ-GAN) settings.}
\label{tab:VQ-GAN}
\tabcolsep=0.1cm
\begin{tabular}{p{1.8cm}p{1.5cm}p{1.8cm}p{0.5cm}p{0.8cm}p{0.8cm}}
\toprule
Model     & Dataset    & $R\to D$ & Rate                    & SSIM   & FID   \\
\midrule
VQ-VAE     & ImageNet   & $256^2\to 16^2$   &   F16                   & 0.7026 & 13.3 \\
VQ-GAN     & ImageNet   & $256^2\to 16^2$   &   F16                   & 0.7105 & 6.04  \\
VQ-GAN     & ImageNet   & $256^2\to 32^2$   &   F8                   & 0.8285 & 2.03  \\
VQ-GAN     & ImageNet   & $336^2\to 21^2$   &    F16                 & 0.7213 & 4.79  \\
VQ-GAN     & OpenImages & $336^2\to 21^2$   &    F16                  & 0.7527 & 4.31  \\
\midrule
Model     & Dataset    & $R\to D$ & Rate   & PA     & FWIoU \\
\midrule
VQ-GAN-Seg & MSCOCO     & $336^2\to 21^2$   &  F16                  & 96.82  & 93.91 \\
VQ-GAN-Seg & VSPW       & $336^2\to 21^2$   &  F16                 & 95.36  & 91.82 \\
\bottomrule
\end{tabular}
\end{center}
	\vspace{-2mm}
\end{table}


The below part of Tab.~\ref{tab:VQ-GAN} shows the performance of VQ-GAN for sketches. VQ-GAN-Seg on MSCOCO\cite{linMicrosoftCocoCommon2014} is trained for Sketch-to-Image (S2I) task and VQ-GAN-Seg on VSPW\cite{miaoVSPWLargescaleDataset2021} is trained for Sketch-to-Video (S2V) task. All the above backbone shows good performance in Pixel Accuracy (PA) and Frequency Weighted Intersection over Union (FWIoU), which shows a good quality of 3D sketch representation used in our model. Fig.~\ref{fig:VQ-GANSketch} also shows some reconstructed samples of 336×336 images and sketches.


Tab.~\ref{tab:pipe} shows the effectiveness of multi-task pre-training for the Text-to-Video (T2V) generation task. We study on a challenging dataset, MSR-VTT\cite{xuMsrvttLargeVideo2016}, with natural descriptions and real-world videos. Compared with training only on a single T2V task (Row 1), training on both T2V and T2I (Row 2) improves the CLIPSIM from 0.2314 to 0.2379. This is because T2I helps to build a connection between text and image, and thus helpful for the semantic consistency of the T2V task. In contrast, training on both T2V and V2V (Row 3) improves the FVD score from 52.98 to 51.81. This is because V2V helps to learn a common unconditional video pattern, and is thus helpful for the visual quality of the T2V task. As a default setting of NÜWA, training on all three tasks achieves the best performance.

Tab.~\ref{tab:sparse} shows the effectiveness of 3D nearby attention for the Sketch-to-Video (S2V) task on the VSPW\cite{miaoVSPWLargescaleDataset2021} dataset. We study on the S2V task because both the encoder and decoder of this task are fed with 3D video data. To evaluate the semantic consistency for S2V, we propose a new metric called Detected PA, which uses a semantic segmentation model\cite{yuanObjectcontextualRepresentationsSemantic2020} to segment each frame of the generated video and then calculate the pixel accuracy between the generated segments and input video sketch. The default NÜWA setting in the last row, with both nearby encoder and nearby decoder, achieves the best FID-vid and Detected PA. The performance drops if either encoder or decoder is replaced by full attention, showing that focusing on nearby conditions and nearby generated results is better than simply considering all the information.
We compare nearby-sparse and axial-sparse in two-folds. Firstly, the computational complexity of nearby-sparse is $O\left(\left(hws\right)\left(e^he^we^s\right)\right)$ and axis-sparse attention is $O\left(\left(hws\right)\left(h+w+s\right)\right)$. For generating long videos (larger $s$), nearby-sparse will be more computational efficient. Secondly, nearby-sparse has better performance than axis-sparse in visual generation task, which is because nearby-sparse attends to ``nearby'' locations containing interactions between both spatial and temporal axes, while axis-sparse handles different axis separately and only consider interactions on the same axis.

\begin{table}[t]
\footnotesize
\begin{center}
\caption{Effectiveness of multi-task pre-training for Text-to-Video (T2V) generation task on MSRVTT dataset.}
\label{tab:pipe}
\begin{tabular}{p{2cm}p{2cm}p{1.5cm}p{1.2cm}}
\toprule
Model  & Pre-trained Tasks & FID-vid↓  & CLIPSIM↑ \\
\midrule
NÜWA-TV   & T2V    &  52.98   &      0.2314      \\
NÜWA-TV-TI  & T2V+T2I    &  53.92   &  0.2379     \\    
NÜWA-TV-VV   & T2V+V2V    &  51.81   &  0.2335     \\    
NÜWA   & T2V+T2I+V2V   &    \textbf{47.68} &  \textbf{0.2439}  \\    
\bottomrule
\end{tabular}
\end{center}
\vspace{-6mm}
\end{table}
\begin{table}[t]
\footnotesize
\begin{center}
\caption{Effectiveness of 3D nearby attention for Sketch-to-Video (S2V) task on VSPW dataset.}
\label{tab:sparse}
\begin{tabular}{p{1.5cm}p{0.9cm}p{0.9cm}p{1.4cm}p{1.6cm}}
\toprule
Model         & Encoder & Decoder & FID-vid↓ & Detected PA↑ \\
\midrule
NÜWA-FF    & Full & Full &35.21 &   0.5220         \\
NÜWA-NF    & Nearby & Full &33.63 &   0.5357         \\
NÜWA-FN & Full & Nearby & 32.06 &    0.5438          \\
NÜWA-AA    & Axis & Axis &29.18  &  0.5957          \\
NÜWA          & Nearby & Nearby &\textbf{27.79} &   \textbf{0.6085}     \\
\bottomrule
\end{tabular}
\end{center}
\vspace{-6mm}
\end{table}

\begin{figure}[t]
	\centering
	\includegraphics[width=3.4in]{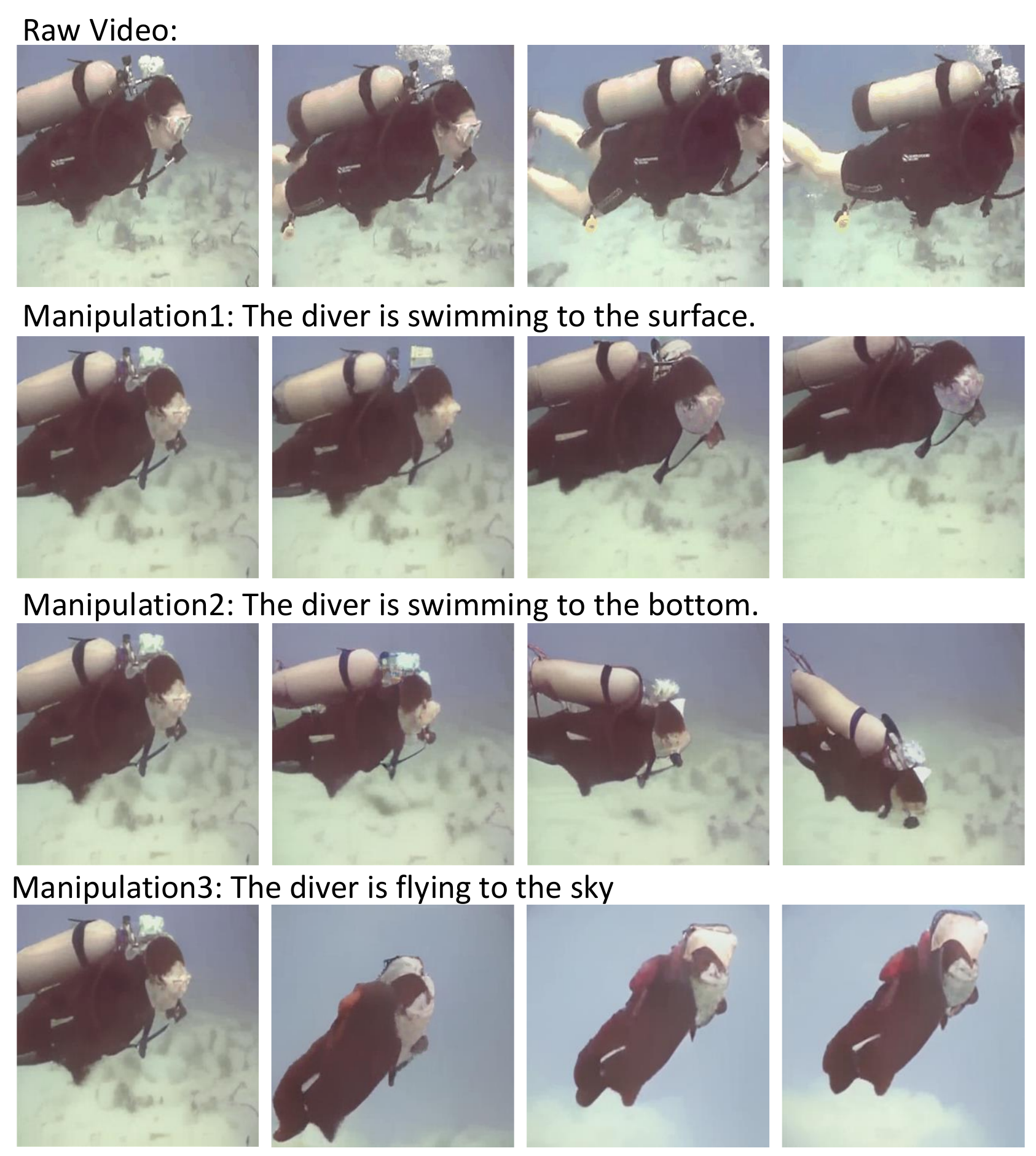}
	\caption{Samples of different manipulations on the same video.}
	\label{fig:TV2V}
	\vspace{-6mm}
\end{figure}

Fig.~\ref{fig:TV2V} shows a new task proposed in this paper, which we call ``Text-Guided Video Manipulation (TV2V)''. TV2V aims to change the future of a video starting from a selected frame guided by text. All samples start to change the future of the video from the second frame. 
The first row shows the original video frames, where a diver is swimming in the water. 
After feeding ``The diver is swimming to the surface'' into NÜWA's encoder and providing the first video frame, NÜWA successfully generates a video with the diver swimming to the surface in the second row. 
The third row shows another successful sample that lets the diver swim to the bottom. What if we want the diver flying to the sky? The fourth row shows that NÜWA can make it as well, where the diver is flying upward, like a rocket.

\section{Conclusion}
In this paper, we present NÜWA as a unified pre-trained model that can generate new or manipulate existing images and videos for 8 visual synthesis tasks. Several contributions are made here, including (1) a general 3D encoder-decoder framework covering texts, images, and videos at the same time; (2) a nearby-sparse attention mechanism that considers the nearby characteristic of both spatial and temporal axes; (3) comprehensive experiments on 8 synthesis tasks. This is our first step towards building an AI platform to enable visual world creation and help content creators.
{\small
\bibliographystyle{ieee_fullname}
\bibliography{wcf}

\begin{thebibliography}{10}\itemsep=-1pt

\bibitem{babaeizadehStochasticVariationalVideo2017a}
Mohammad Babaeizadeh, Chelsea Finn, Dumitru Erhan, Roy~H. Campbell, and Sergey
  Levine.
\newblock Stochastic variational video prediction.
\newblock {\em arXiv preprint arXiv:1710.11252}, 2017.

\bibitem{balajiConditionalGANDiscriminative2019}
Yogesh Balaji, Martin~Renqiang Min, Bing Bai, Rama Chellappa, and Hans~Peter
  Graf.
\newblock Conditional {{GAN}} with {{Discriminative Filter Generation}} for
  {{Text}}-to-{{Video Synthesis}}.
\newblock In {\em {{IJCAI}}}, pages 1995--2001, 2019.

\bibitem{bauPaintWord2021}
David Bau, Alex Andonian, Audrey Cui, YeonHwan Park, Ali Jahanian, Aude Oliva,
  and Antonio Torralba.
\newblock Paint by word.
\newblock {\em arXiv preprint arXiv:2103.10951}, 2021.

\bibitem{brockLargeScaleGAN2019}
Andrew Brock, Jeff Donahue, and Karen Simonyan.
\newblock Large {{Scale GAN Training}} for {{High Fidelity Natural Image
  Synthesis}}.
\newblock {\em arXiv:1809.11096 [cs, stat]}, Feb. 2019.

\bibitem{chenGenerativepretrainingPixels2020}
Mark Chen, Alec Radford, Rewon Child, Jeffrey Wu, Heewoo Jun, David Luan, and
  Ilya Sutskever.
\newblock Generative pretraining from pixels.
\newblock In {\em International {{Conference}} on {{Machine Learning}}}, pages
  1691--1703. {PMLR}, 2020.

\bibitem{childGeneratingLongSequences2019}
Rewon Child, Scott Gray, Alec Radford, and Ilya Sutskever.
\newblock Generating long sequences with sparse transformers.
\newblock {\em arXiv preprint arXiv:1904.10509}, 2019.

\bibitem{clarkWhatDoesBert2019}
Kevin Clark, Urvashi Khandelwal, Omer Levy, and Christopher~D. Manning.
\newblock What does bert look at? an analysis of bert's attention.
\newblock {\em arXiv preprint arXiv:1906.04341}, 2019.

\bibitem{dentonStochasticVideoGeneration2018}
Emily Denton and Rob Fergus.
\newblock Stochastic video generation with a learned prior.
\newblock In {\em International {{Conference}} on {{Machine Learning}}}, pages
  1174--1183. {PMLR}, 2018.

\bibitem{dingCogViewMasteringTexttoImage2021}
Ming Ding, Zhuoyi Yang, Wenyi Hong, Wendi Zheng, Chang Zhou, Da Yin, Junyang
  Lin, Xu Zou, Zhou Shao, Hongxia Yang, and Jie Tang.
\newblock {{CogView}}: Mastering {{Text}}-to-{{Image Generation}} via
  {{Transformers}}.
\newblock {\em arXiv:2105.13290 [cs]}, May 2021.

\bibitem{ebertSelfSupervisedVisualPlanning2017}
Frederik Ebert, Chelsea Finn, Alex~X. Lee, and Sergey Levine.
\newblock Self-{{Supervised Visual Planning}} with {{Temporal Skip
  Connections}}.
\newblock In {\em {{CoRL}}}, pages 344--356, 2017.

\bibitem{esserTamingTransformersHighResolution2021}
Patrick Esser, Robin Rombach, and Bj{\"o}rn Ommer.
\newblock Taming {{Transformers}} for {{High}}-{{Resolution Image Synthesis}}.
\newblock {\em arXiv:2012.09841 [cs]}, June 2021.

\bibitem{finnUnsupervisedLearningPhysical2016a}
Chelsea Finn, Ian Goodfellow, and Sergey Levine.
\newblock Unsupervised learning for physical interaction through video
  prediction.
\newblock {\em Advances in neural information processing systems}, 29:64--72,
  2016.

\bibitem{franceschiStochasticLatentResidual2020}
Jean-Yves Franceschi, Edouard Delasalles, Micka{\"e}l Chen, Sylvain Lamprier,
  and Patrick Gallinari.
\newblock Stochastic latent residual video prediction.
\newblock In {\em International {{Conference}} on {{Machine Learning}}}, pages
  3233--3246. {PMLR}, 2020.

\bibitem{heuselGansTrainedTwo2017}
Martin Heusel, Hubert Ramsauer, Thomas Unterthiner, Bernhard Nessler, and Sepp
  Hochreiter.
\newblock Gans trained by a two time-scale update rule converge to a local nash
  equilibrium.
\newblock {\em Advances in neural information processing systems}, 30, 2017.

\bibitem{hoAxialAttentionMultidimensional2019}
Jonathan Ho, Nal Kalchbrenner, Dirk Weissenborn, and Tim Salimans.
\newblock Axial {{Attention}} in {{Multidimensional Transformers}}.
\newblock {\em arXiv preprint arXiv:1912.12180}, 2019.

\bibitem{kayKineticsHumanAction2017}
Will Kay, Joao Carreira, Karen Simonyan, Brian Zhang, Chloe Hillier, Sudheendra
  Vijayanarasimhan, Fabio Viola, Tim Green, Trevor Back, and Paul Natsev.
\newblock The kinetics human action video dataset.
\newblock {\em arXiv preprint arXiv:1705.06950}, 2017.

\bibitem{Kingma_Adammethodstochastic_2014}
Diederik Kingma and Jimmy Ba.
\newblock Adam: A method for stochastic optimization.
\newblock {\em arXiv preprint arXiv:1412.6980}, 2014.

\bibitem{kumarVideoflowConditionalFlowbased2019}
Manoj Kumar, Mohammad Babaeizadeh, Dumitru Erhan, Chelsea Finn, Sergey Levine,
  Laurent Dinh, and Durk Kingma.
\newblock Videoflow: A conditional flow-based model for stochastic video
  generation.
\newblock {\em arXiv preprint arXiv:1903.01434}, 2019.

\bibitem{kuznetsovaOpenImagesDataset2020}
Alina Kuznetsova, Hassan Rom, Neil Alldrin, Jasper Uijlings, Ivan Krasin, Jordi
  {Pont-Tuset}, Shahab Kamali, Stefan Popov, Matteo Malloci, and Alexander
  Kolesnikov.
\newblock The open images dataset v4.
\newblock {\em International Journal of Computer Vision}, 128(7):1956--1981,
  2020.

\bibitem{leeStochasticAdversarialVideo2018}
Alex~X. Lee, Richard Zhang, Frederik Ebert, Pieter Abbeel, Chelsea Finn, and
  Sergey Levine.
\newblock Stochastic adversarial video prediction.
\newblock {\em arXiv preprint arXiv:1804.01523}, 2018.

\bibitem{liVideoGenerationText2018}
Yitong Li, Martin Min, Dinghan Shen, David Carlson, and Lawrence Carin.
\newblock Video generation from text.
\newblock In {\em Proceedings of the {{AAAI Conference}} on {{Artificial
  Intelligence}}}, volume~32, 2018.

\bibitem{linMicrosoftCocoCommon2014}
Tsung-Yi Lin, Michael Maire, Serge Belongie, James Hays, Pietro Perona, Deva
  Ramanan, Piotr Doll{\'a}r, and C.~Lawrence Zitnick.
\newblock Microsoft coco: Common objects in context.
\newblock In {\em European Conference on Computer Vision}, pages 740--755.
  {Springer}, 2014.

\bibitem{lucTransformationbasedAdversarialVideo2020}
Pauline Luc, Aidan Clark, Sander Dieleman, Diego de~Las Casas, Yotam Doron,
  Albin Cassirer, and Karen Simonyan.
\newblock Transformation-based adversarial video prediction on large-scale
  data.
\newblock {\em arXiv preprint arXiv:2003.04035}, 2020.

\bibitem{miaoVSPWLargescaleDataset2021}
Jiaxu Miao, Yunchao Wei, Yu Wu, Chen Liang, Guangrui Li, and Yi Yang.
\newblock {{VSPW}}: A {{Large}}-scale {{Dataset}} for {{Video Scene Parsing}}
  in the {{Wild}}.
\newblock In {\em Proceedings of the {{IEEE}}/{{CVF Conference}} on {{Computer
  Vision}} and {{Pattern Recognition}}}, pages 4133--4143, 2021.

\bibitem{moingCCVSContextawareControllable2021}
Guillaume~Le Moing, Jean Ponce, and Cordelia Schmid.
\newblock {{CCVS}}: Context-aware {{Controllable Video Synthesis}}.
\newblock {\em arXiv preprint arXiv:2107.08037}, 2021.

\bibitem{monfortMomentsTimeDataset2019}
Mathew Monfort, Alex Andonian, Bolei Zhou, Kandan Ramakrishnan, Sarah~Adel
  Bargal, Tom Yan, Lisa Brown, Quanfu Fan, Dan Gutfreund, and Carl Vondrick.
\newblock Moments in time dataset: One million videos for event understanding.
\newblock {\em IEEE transactions on pattern analysis and machine intelligence},
  42(2):502--508, 2019.

\bibitem{parkSemanticImageSynthesis2019}
Taesung Park, Ming-Yu Liu, Ting-Chun Wang, and Jun-Yan Zhu.
\newblock Semantic image synthesis with spatially-adaptive normalization.
\newblock In {\em Proceedings of the {{IEEE}}/{{CVF Conference}} on {{Computer
  Vision}} and {{Pattern Recognition}}}, pages 2337--2346, 2019.

\bibitem{parmarImageTransformer2018}
Niki Parmar, Ashish Vaswani, Jakob Uszkoreit, {\textbackslash}Lukasz Kaiser,
  Noam Shazeer, Alexander Ku, and Dustin Tran.
\newblock Image transformer.
\newblock {\em arXiv preprint arXiv:1802.05751}, 2018.

\bibitem{radfordLearningTransferableVisual2021}
Alec Radford, Jong~Wook Kim, Chris Hallacy, Aditya Ramesh, Gabriel Goh,
  Sandhini Agarwal, Girish Sastry, Amanda Askell, Pamela Mishkin, Jack Clark,
  Gretchen Krueger, and Ilya Sutskever.
\newblock Learning {{Transferable Visual Models From Natural Language
  Supervision}}.
\newblock {\em arXiv:2103.00020 [cs]}, Feb. 2021.

\bibitem{radfordUnsupervisedRepresentationLearning2015}
Alec Radford, Luke Metz, and Soumith Chintala.
\newblock Unsupervised representation learning with deep convolutional
  generative adversarial networks.
\newblock {\em arXiv preprint arXiv:1511.06434}, 2015.

\bibitem{rakhimovLatentVideoTransformer2020}
Ruslan Rakhimov, Denis Volkhonskiy, Alexey Artemov, Denis Zorin, and Evgeny
  Burnaev.
\newblock Latent {{Video Transformer}}.
\newblock {\em arXiv preprint arXiv:2006.10704}, 2020.

\bibitem{ramachandranStandaloneSelfattentionVision2019}
Prajit Ramachandran, Niki Parmar, Ashish Vaswani, Irwan Bello, Anselm Levskaya,
  and Jonathon Shlens.
\newblock Stand-alone self-attention in vision models.
\newblock {\em arXiv preprint arXiv:1906.05909}, 2019.

\bibitem{rameshZeroShotTexttoImageGeneration2021}
Aditya Ramesh, Mikhail Pavlov, Gabriel Goh, Scott Gray, Chelsea Voss, Alec
  Radford, Mark Chen, and Ilya Sutskever.
\newblock Zero-{{Shot Text}}-to-{{Image Generation}}.
\newblock {\em arXiv:2102.12092 [cs]}, Feb. 2021.

\bibitem{russakovskyImageNetLargeScale2015}
Olga Russakovsky, Jia Deng, Hao Su, Jonathan Krause, Sanjeev Satheesh, Sean Ma,
  Zhiheng Huang, Andrej Karpathy, Aditya Khosla, Michael Bernstein,
  Alexander~C. Berg, and Li {Fei-Fei}.
\newblock {{ImageNet Large Scale Visual Recognition Challenge}}.
\newblock {\em arXiv:1409.0575 [cs]}, Jan. 2015.

\bibitem{salimansImprovedTechniquesTraining2016}
Tim Salimans, Ian Goodfellow, Wojciech Zaremba, Vicki Cheung, Alec Radford, and
  Xi Chen.
\newblock Improved techniques for training gans.
\newblock {\em Advances in neural information processing systems},
  29:2234--2242, 2016.

\bibitem{taoDfganDeepFusion2020}
Ming Tao, Hao Tang, Songsong Wu, Nicu Sebe, Xiao-Yuan Jing, Fei Wu, and Bingkun
  Bao.
\newblock Df-gan: Deep fusion generative adversarial networks for text-to-image
  synthesis.
\newblock {\em arXiv preprint arXiv:2008.05865}, 2020.

\bibitem{tulyakovMocoganDecomposingMotion2018}
Sergey Tulyakov, Ming-Yu Liu, Xiaodong Yang, and Jan Kautz.
\newblock Mocogan: Decomposing motion and content for video generation.
\newblock In {\em Proceedings of the {{IEEE}} Conference on Computer Vision and
  Pattern Recognition}, pages 1526--1535, 2018.

\bibitem{unterthinerAccurateGenerativeModels2018}
Thomas Unterthiner, Sjoerd {van Steenkiste}, Karol Kurach, Raphael Marinier,
  Marcin Michalski, and Sylvain Gelly.
\newblock Towards accurate generative models of video: A new metric \&
  challenges.
\newblock {\em arXiv preprint arXiv:1812.01717}, 2018.

\bibitem{oordConditionalImageGeneration2016}
Aaron van~den Oord, Nal Kalchbrenner, Oriol Vinyals, Lasse Espeholt, Alex
  Graves, and Koray Kavukcuoglu.
\newblock Conditional image generation with pixelcnn decoders.
\newblock {\em arXiv preprint arXiv:1606.05328}, 2016.

\bibitem{oordNeuralDiscreteRepresentation2017}
Aaron van~den Oord, Oriol Vinyals, and Koray Kavukcuoglu.
\newblock Neural discrete representation learning.
\newblock {\em arXiv preprint arXiv:1711.00937}, 2017.

\bibitem{vanoordPixelRecurrentNeural2016}
Aaron Van~Oord, Nal Kalchbrenner, and Koray Kavukcuoglu.
\newblock Pixel recurrent neural networks.
\newblock In {\em International {{Conference}} on {{Machine Learning}}}, pages
  1747--1756. {PMLR}, 2016.

\bibitem{vaswaniAttentionAllYou2017}
Ashish Vaswani, Noam Shazeer, Niki Parmar, Jakob Uszkoreit, Llion Jones,
  Aidan~N. Gomez, {\textbackslash}Lukasz Kaiser, and Illia Polosukhin.
\newblock Attention is all you need.
\newblock In {\em Advances in Neural Information Processing Systems}, pages
  5998--6008, 2017.

\bibitem{wangVatexLargescaleHighquality2019}
Xin Wang, Jiawei Wu, Junkun Chen, Lei Li, Yuan-Fang Wang, and William~Yang
  Wang.
\newblock Vatex: A large-scale, high-quality multilingual dataset for
  video-and-language research.
\newblock In {\em Proceedings of the {{IEEE}}/{{CVF International Conference}}
  on {{Computer Vision}}}, pages 4581--4591, 2019.

\bibitem{weissenbornScalingAutoregressiveVideo2020}
Dirk Weissenborn, Oscar T{\"a}ckstr{\"o}m, and Jakob Uszkoreit.
\newblock Scaling autoregressive video models.
\newblock In {\em {{ICLR}}}, 2020.

\bibitem{wuGODIVAGeneratingOpenDomaIn2021}
Chenfei Wu, Lun Huang, Qianxi Zhang, Binyang Li, Lei Ji, Fan Yang, Guillermo
  Sapiro, and Nan Duan.
\newblock {{GODIVA}}: Generating {{Open}}-{{DomaIn Videos}} from {{nAtural
  Descriptions}}.
\newblock {\em arXiv:2104.14806 [cs]}, Apr. 2021.

\bibitem{xuMsrvttLargeVideo2016}
Jun Xu, Tao Mei, Ting Yao, and Yong Rui.
\newblock Msr-vtt: A large video description dataset for bridging video and
  language.
\newblock In {\em Proceedings of the {{IEEE}} Conference on Computer Vision and
  Pattern Recognition}, pages 5288--5296, 2016.

\bibitem{xuAttnganFinegrainedText2018}
Tao Xu, Pengchuan Zhang, Qiuyuan Huang, Han Zhang, Zhe Gan, Xiaolei Huang, and
  Xiaodong He.
\newblock Attngan: Fine-grained text to image generation with attentional
  generative adversarial networks.
\newblock In {\em Proceedings of the {{IEEE}} Conference on Computer Vision and
  Pattern Recognition}, pages 1316--1324, 2018.

\bibitem{yanVideoGPTVideoGeneration2021}
Wilson Yan, Yunzhi Zhang, Pieter Abbeel, and Aravind Srinivas.
\newblock {{VideoGPT}}: Video {{Generation}} using {{VQ}}-{{VAE}} and
  {{Transformers}}.
\newblock {\em arXiv preprint arXiv:2104.10157}, 2021.

\bibitem{yuanObjectcontextualRepresentationsSemantic2020}
Yuhui Yuan, Xilin Chen, and Jingdong Wang.
\newblock Object-contextual representations for semantic segmentation.
\newblock In {\em Computer {{Vision}}\textendash{{ECCV}} 2020: 16th {{European
  Conference}}, {{Glasgow}}, {{UK}}, {{August}} 23\textendash 28, 2020,
  {{Proceedings}}, {{Part VI}} 16}, pages 173--190. {Springer}, 2020.

\bibitem{zhangCrossmodalContrastiveLearning2021}
Han Zhang, Jing~Yu Koh, Jason Baldridge, Honglak Lee, and Yinfei Yang.
\newblock Cross-modal contrastive learning for text-to-image generation.
\newblock In {\em Proceedings of the {{IEEE}}/{{CVF Conference}} on {{Computer
  Vision}} and {{Pattern Recognition}}}, pages 833--842, 2021.

\bibitem{zhangVideoGenGenerativeModeling2020}
Yunzhi Zhang, Wilson Yan, Pieter Abbeel, and Aravind Srinivas.
\newblock {{VideoGen}}: Generative {{Modeling}} of {{Videos}} using
  {{VQ}}-{{VAE}} and {{Transformers}}.
\newblock Sept. 2020.

\bibitem{zhuDmganDynamicMemory2019}
Minfeng Zhu, Pingbo Pan, Wei Chen, and Yi Yang.
\newblock Dm-gan: Dynamic memory generative adversarial networks for
  text-to-image synthesis.
\newblock In {\em Proceedings of the {{IEEE}}/{{CVF Conference}} on {{Computer
  Vision}} and {{Pattern Recognition}}}, pages 5802--5810, 2019.

\end{thebibliography}
}

\twocolumn[{
\renewcommand\twocolumn[1][]{#1}
\maketitle
\begin{center}
    \centering
    \captionsetup{type=figure}
    \includegraphics[width=\textwidth]{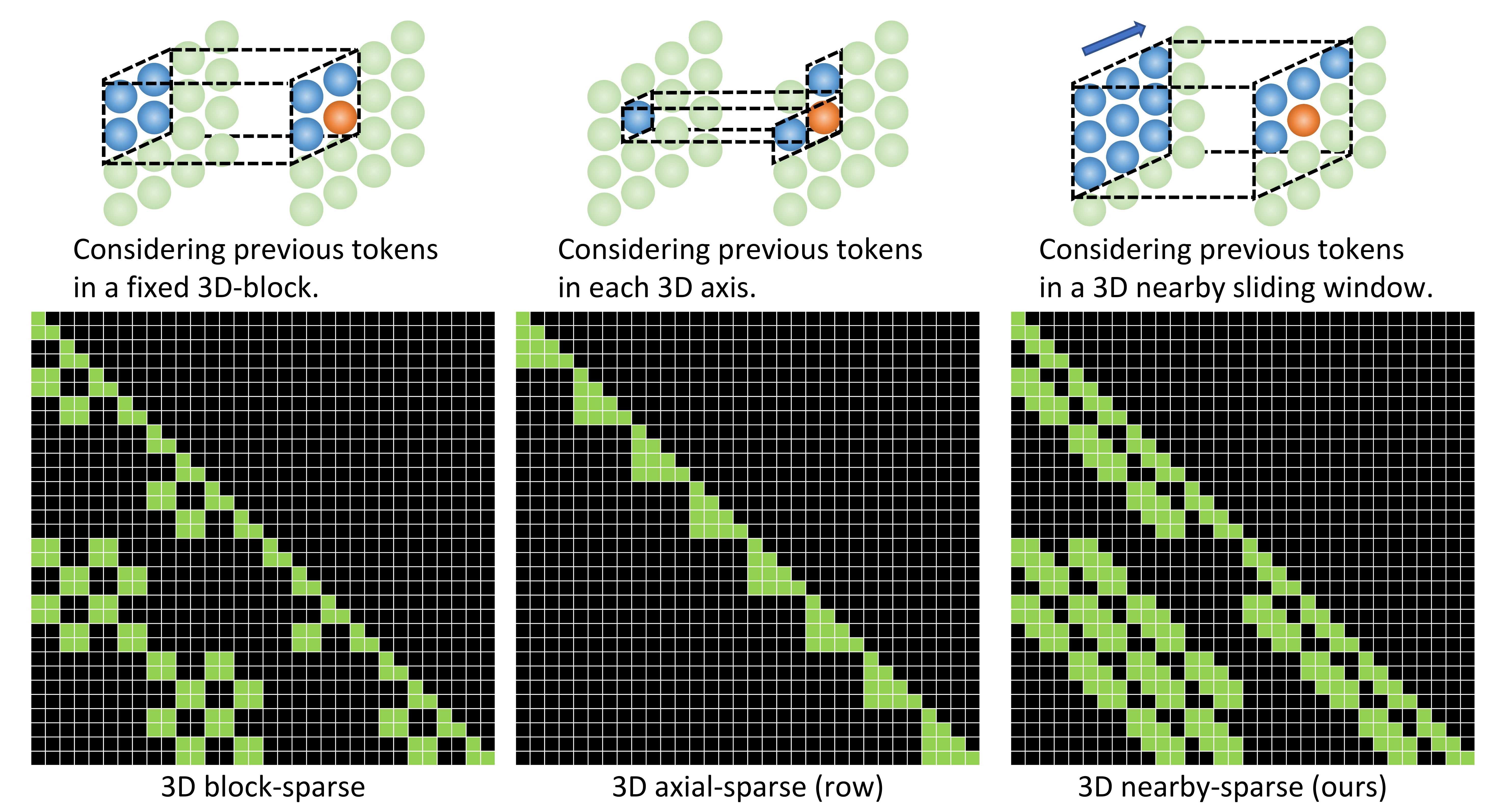}
    \captionof{figure}{Comparisons between different 3D sparse attentions. All samples assume that the size of the input 3D data is  $4\times 4\times 2=32$. The illustrations in the upper part show which tokens (blue) need to be attended to generate the target token (orange). The matrices of the size $32\times 32$ in the lower part show the attention masks in sparse attention (black denotes masked tokens).}
    \label{fig:Sparse}
\end{center}
}]
\thispagestyle{empty}
\appendix

\section{Comparisons between 3D Sparse Attentions}

\begin{table}[t]
\footnotesize
\begin{center}
\caption{Complexity of different 3D sparse attention.}
\label{tab:sparse}
\begin{tabular}{p{3.6cm}p{3.5cm}}
\toprule
Module         & Complexity \\
\midrule
3D full  & $O\left((hws)^2\right)$      \\
3D block-sparse~\cite{weissenbornScalingAutoregressiveVideo2020, rakhimovLatentVideoTransformer2020}     & $O\left(\left(\frac{hws}{b}\right)^2\right)$      \\
3D axial-sparse~\cite{hoAxialAttentionMultidimensional2019, rameshZeroShotTexttoImageGeneration2021, wuGODIVAGeneratingOpenDomaIn2021}   & $O\left(\left(hws\right)\left(h+w+s\right)\right)$   \\
3D nearby-sparse (ours)    & $O\left(\left(hws\right)\left(e^he^we^s\right)\right)$      \\
\bottomrule
\end{tabular}
\end{center}
\end{table}

Fig.~\ref{fig:Sparse} shows comparisons between different 3D sparse attentions. Assume we have 3D data with the size of $4\times 4\times 2$, the idea of 3D block-sparse attention is to split the 3D data into several fixed blocks and handle these blocks separately. There are many ways to split blocks, such as splitting in time, space, or both. The 3D block-sparse example in Fig.~\ref{fig:Sparse} considers the split of both time and space. The 3D data is divided into 4 parts, each has the size of $2\times 2 \times 2$. To generate the orange token, 3D block-sparse attention considers previous tokens inside the fixed 3D block. Although 3D block-sparse attention considers both spatial and temporal axes, this spatial and temporal information is limited and fixed in the 3D block especially for the tokens along the edge of the 3D block. Only part of nearby information is considered since some nearby information outside the 3D block is invisible for tokens inside it. The idea of 3D axial-sparse attention is to consider previous tokens along the axis. Although 3D axis-sparse attention considers both spatial and temporal axes, this spatial and temporal information is limited along the axes. Only part of nearby information is considered and some nearby information that does not in the axis will not be considered in the 3D axis attention. In this paper, we propose a 3D nearby-sparse, which considers the full nearby information and dynamically generates the 3D nearby attention block for each token. The attention matrix also shows the evidence as the attended part (blue) for 3D nearby-sparse is more smooth than 3D block-sparse and 3D axial-sparse.

Tab.~\ref{tab:sparse} shows the complexity of different 3D sparse attention. $h, w, s$ denotes the spatial height, spatial width, and temporal length of the 3D data. Different sparse mechanisms have their computational advantages in different scenarios. For example, for long videos or high-resolution frames with large $h, w, s$, usually $\left(e^he^we^s\right) < (h+w+s)$, and 3D nearby-sparse attention is more efficient than 3D axial-sparse attention. If the 3D data can be split into several parts without dependencies, 3D block-sparse will be a good choice. For example, a cartoon with several episodes and each tells a separate story, we can simply split these stories as they share no relationship.

\section{Details of Multi-task Pre-training}

\begin{table}[t]
\footnotesize
\caption{Implementation details for two settings of NÜWA.}
\label{tab:detail}
\begin{tabular}{p{3.5cm}p{1.9cm}p{2.0cm}}
\toprule
Settings                                   & NÜWA-256                                   & NÜWA-336                                     \\

\midrule
VQGAN  image resolution                  & 256×256 & 336×336 \\
VQGAN  discrete tokens                  & 32×32 & 21×21\\
VQGAN   Compression Ratio                  & F8                                           & F16                                          \\
VQGAN   codebook dimension                 & 256                                          & 256                                          \\

\midrule
3DNA  hidden size                          & 1280                                         & 1280                                         \\
3DNA   number of heads                     & 20                                           & 20                                           \\
3DNA   dimension for each head           & 64                                           & 64                                           \\
\midrule
NÜWA   Encoder layers                      & 12                                           & 12                                           \\
NÜWA   Decoder layers                      & 24                                           & 24                                           \\
\midrule
\multirow{3}{*}{Multi-task pretraining datasets} & \multicolumn{2}{c}{Conceptual Captions} \\
& \multicolumn{2}{c}{Moments in Time} \\
& \multicolumn{2}{c}{Vatex} \\
\midrule[0.1pt]
\multirow{3}{*}{Multi-task input 3D size}  & 1×1×77 (T2I)                                  & 1×1×77 (T2I)                                  \\
                                           & 32×32×1 (V2V)                                 & 21×21×1 (V2V)                                 \\
                                           & 1×1×77 (T2V)                                  & 1×1×77 (T2V)                                  \\
                                           \midrule[0.1pt]
\multirow{3}{*}{Multi-task output 3D size} & 32×32×1 (T2I)                                 & 21×21×1 (T2I)                                 \\
                                           & 32×32×4 (V2V)                                 & 21×21×10 (V2V)                                \\
                                           & 32×32×4 (T2V)                                 & 21×21×10 (T2V)                                \\
                                           \midrule
Training   batch size                      & \multicolumn{2}{c}{128}                                                                     \\
Training   learning rate                   & \multicolumn{2}{c}{$10^{-3}$}                                              \\
Training   steps                           & \multicolumn{2}{c}{50M}      \\         \bottomrule
                                                    
\end{tabular}
\end{table}

Tab.~\ref{tab:detail} shows the implementation details of two NÜWA settings used in this paper. Both NÜWA-256 and NÜWA-336 models are trade-off between image quality and video length (number of video frames). As the image quality highly relies on compression ratio and number of discrete tokens, and low compression ratio and large discrete tokens are key factors for high quality image. However, as the total capacity of the model is limited, the number of discrete tokens per image and the number of video frames (images) are a compromise.

Note that  NÜWA-256 adopts a compression ratio of F8 and the discrete tokens is 32×32, while NÜWA-336 adopts a compression ratio of F8 and the discrete tokens is only 21×21. To make a fair comparison with current state-of-the-art models, we adopt NÜWA-256 with more discrete tokens to generate high quality images. However, NÜWA-256 can only generate videos with 4 frames considering the efficiency of transformer. To handle relatively long videos, NÜWA-336 with 
fewer discrete tokens can generate videos with 10 frames. As a result, NÜWA-336 significantly relieves the pressure of the auto-regressive models in the second stage, especially for videos. NÜWA-336 is the default setting to cover both images and videos. 


For both models, note that we did not over-adjust the parameters and just use the same learning rate of $10^{-3}$ and 50M training steps.

\begin{figure}[t]
	\centering
	\includegraphics[width=3.4in]{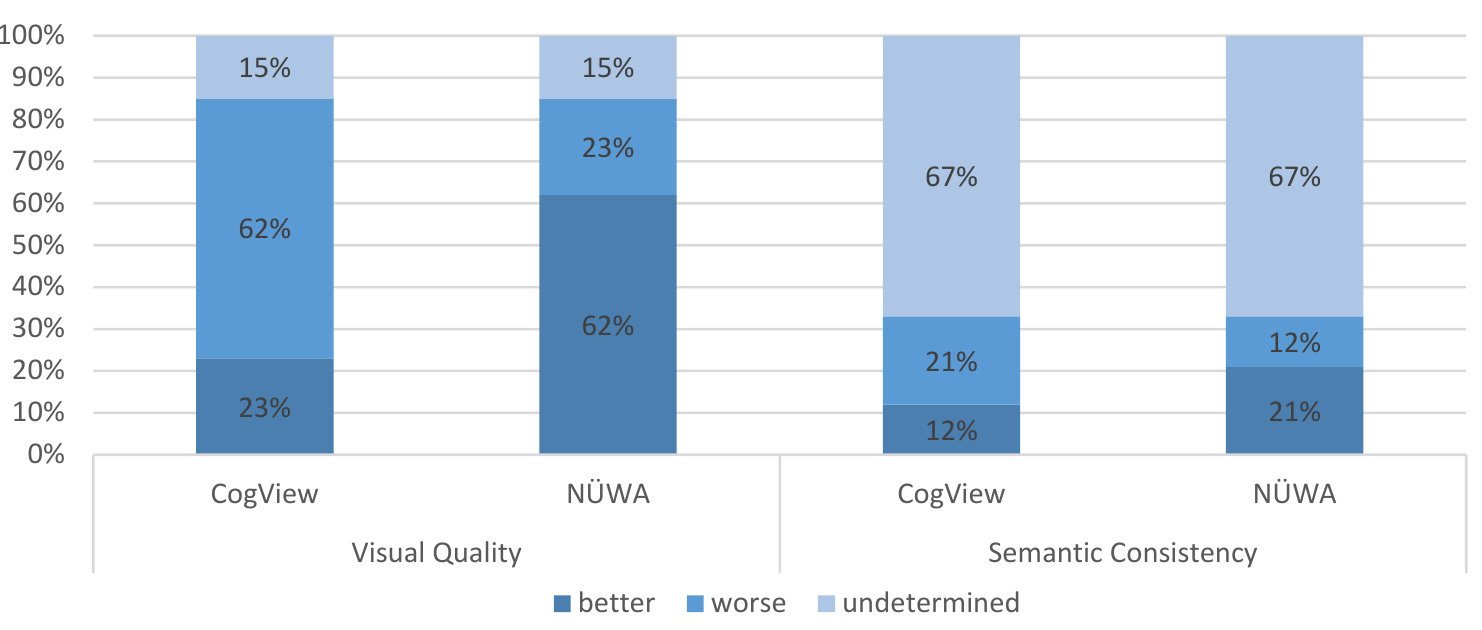}
	\caption{Human evaluation on MSCOCO dataset for Text-to-Image (T2I) task.}
	\label{fig:supp_cogview}
		\vspace{-4mm}

\end{figure}

\begin{figure}[t]
	\centering
	\includegraphics[width=3.4in]{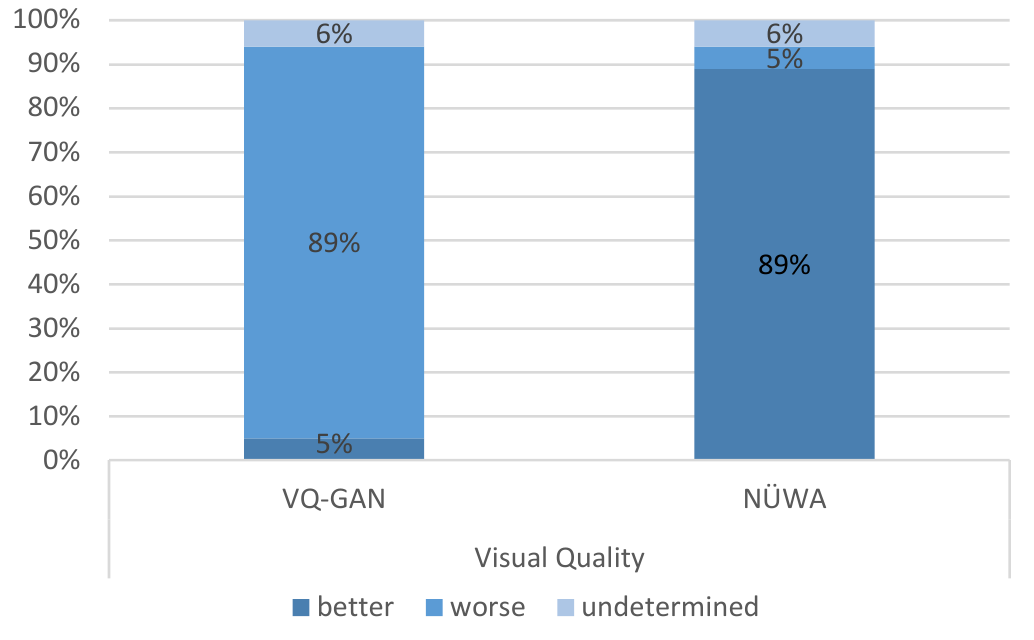}
	\caption{Human evaluation on MSCOCO dataset for Image Completion (I2I) task.}
	\label{fig:supp_vqgan}
	\vspace{-4mm}
\end{figure}
\section{Human Evaluation}
Fig.~\ref{fig:supp_cogview} presents human comparison results between CogView\cite{dingCogViewMasteringTexttoImage2021} and our NÜWA on the MSCOCO dataset for Text-to-Image (T2I) task. We randomly selected 2000 texts and ask annotators to compare the generated results between two models including both visual quality and semantic consistency. The annotators are asked to choose among three options: better, worse, or undetermined. In the visual quality part, There are 62\% votes for our NÜWA model, 15\% undetermined, and 23\% votes for CogView, which shows NÜWA generates more realistic images. In the semantic consistency part, although 67\% of votes cannot determine which model is more consistent with the text, NÜWA also wins the remaining 21\% votes. Although CogView is pretrained on larger text-image pairs than NÜWA, our model still benefits from multi-task pretraining, as text-videos pairs provides high-level semantic information for text-to-image generation.

Fig.~\ref{fig:supp_vqgan} shows human comparison results between VQ-GAN\cite{esserTamingTransformersHighResolution2021} and our NÜWA model on the MSCOCO dataset for the Image Completion (I2I) task. We use similar settings as Fig.~\ref{fig:supp_cogview}, but removed semantic consistency as there is no text input for this task. The comparison results show that there are 89\% votes for NÜWA, which shows the strong zero-shot ability of NÜWA.

\begin{figure*}[h]
	\centering
	\includegraphics[width=\textwidth]{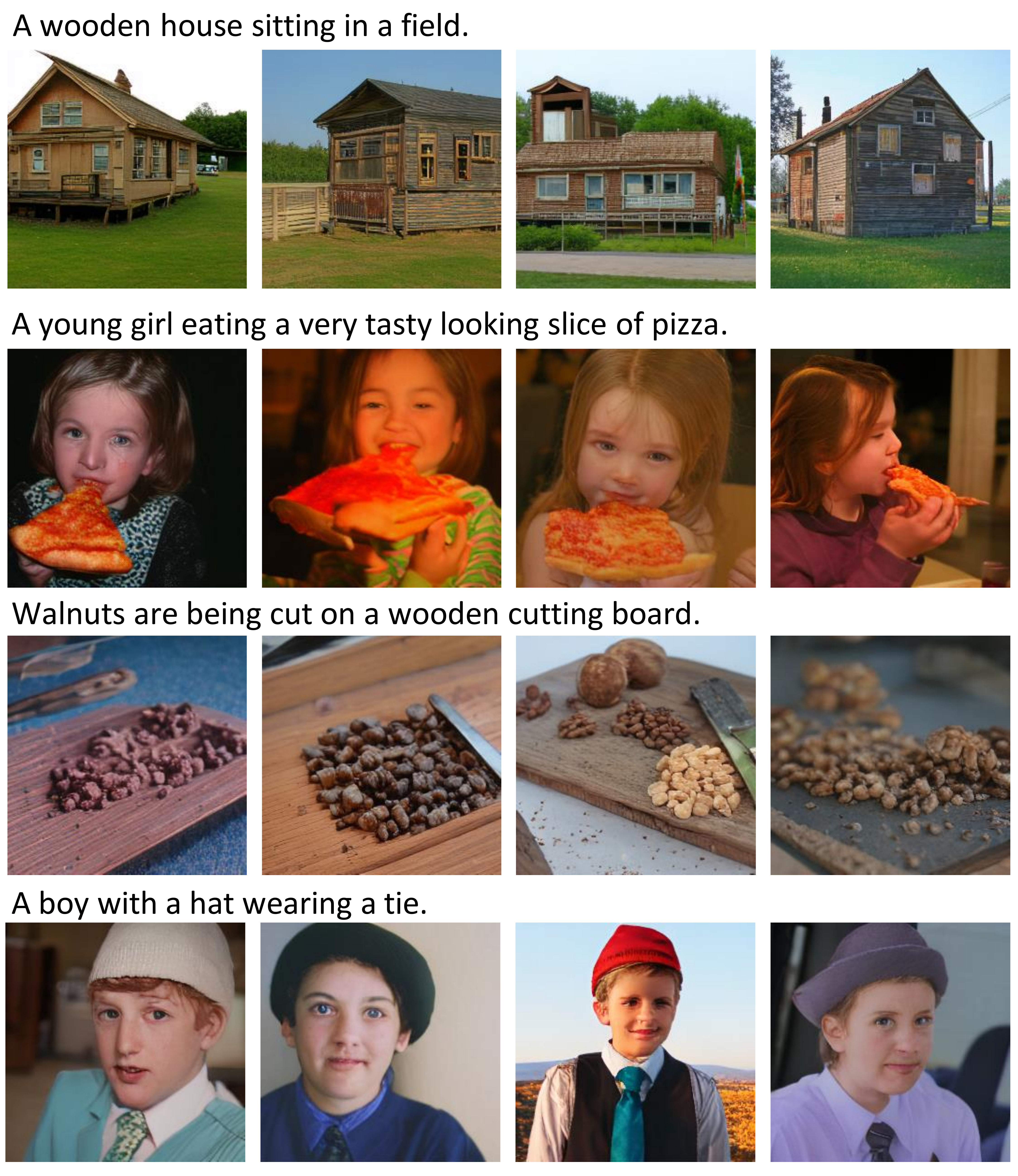}
	\caption{More samples of Text-to-Image (T2I) task generated by NÜWA.}
	\label{fig:supp_t2i_1}
\end{figure*}

\begin{figure*}[h]
	\centering
	\includegraphics[width=\textwidth]{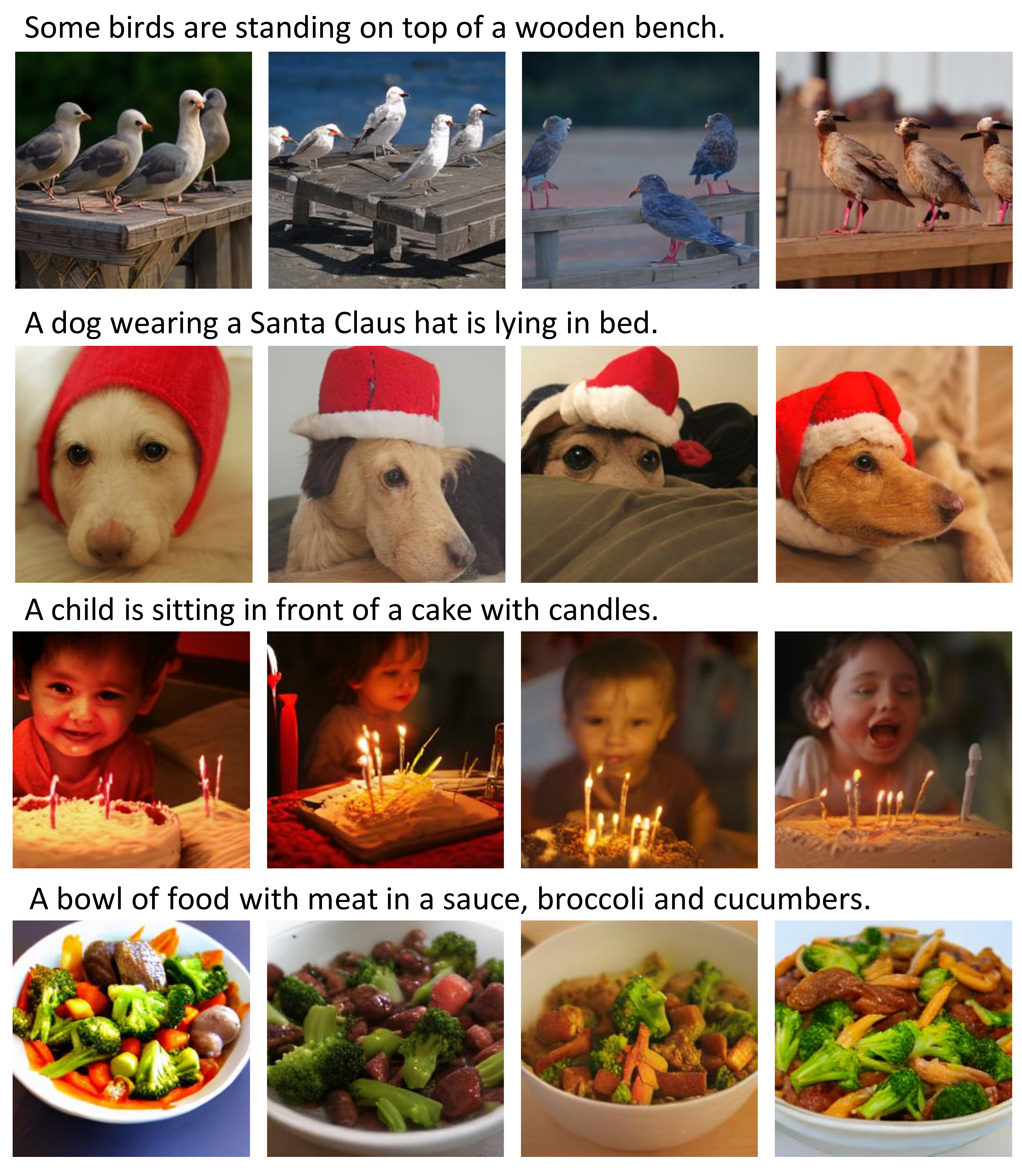}
	\caption{More samples of Text-to-Image (T2I) task generated by NÜWA.}
	\label{fig:supp_t2i_2}
\end{figure*}

\begin{figure*}[h]
	\centering
	\includegraphics[width=\textwidth]{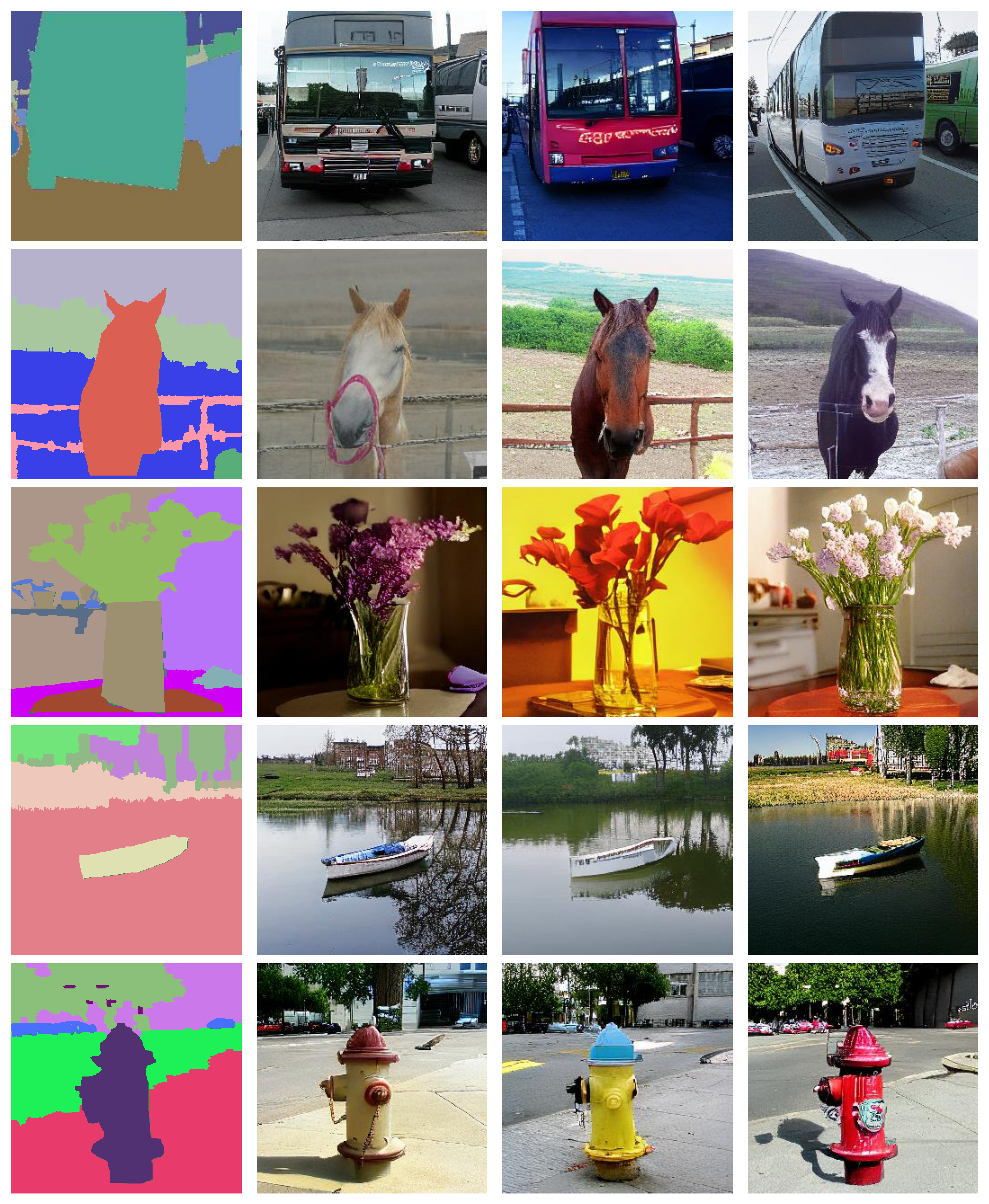}
	\caption{More samples of Sketch-to-Image (S2I) task generated by NÜWA.}
	\label{fig:supp_s2i_1}
\end{figure*}

\begin{figure*}[h]
	\centering
	\includegraphics[width=\textwidth]{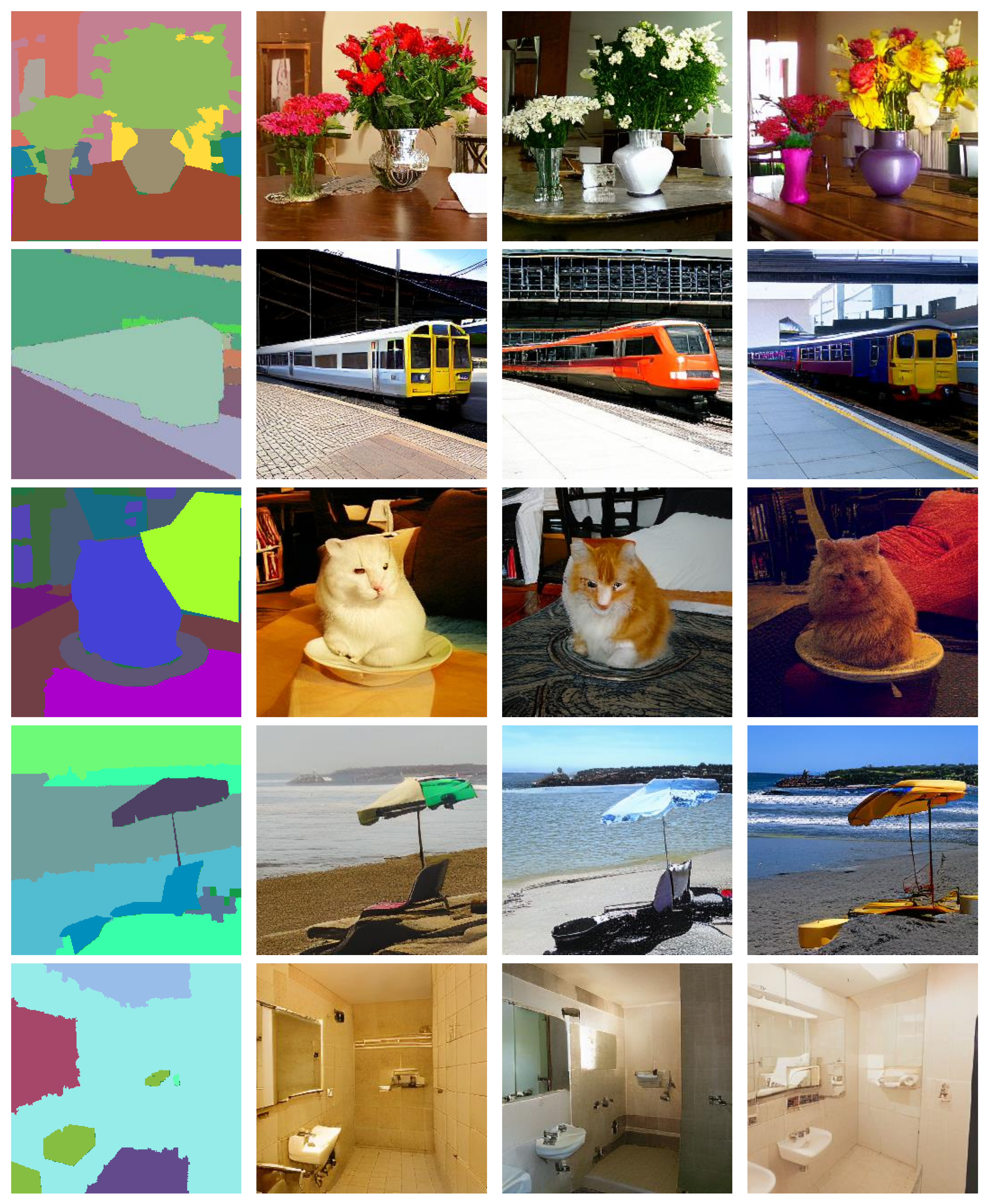}
	\caption{More samples of Sketch-to-Image (S2I) task generated by NÜWA.}
	\label{fig:supp_s2i_2}
\end{figure*}

\begin{figure*}[h]
	\centering
	\includegraphics[width=\textwidth]{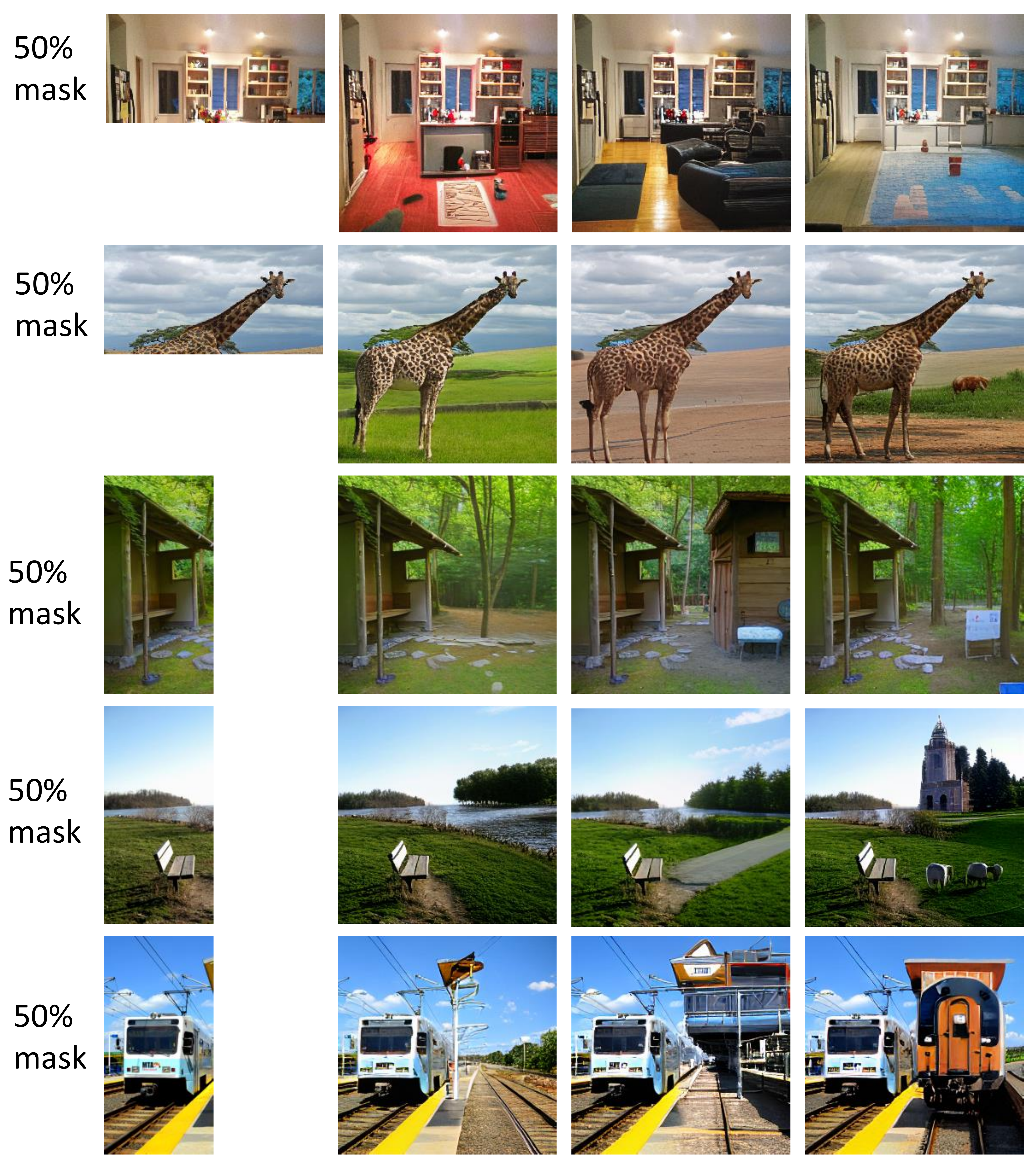}
	\caption{More samples of the Image Completion (I2I) task generated by NÜWA.}
	\label{fig:supp_s2i_1}
\end{figure*}

\begin{figure*}[h]
	\centering
	\includegraphics[width=\textwidth]{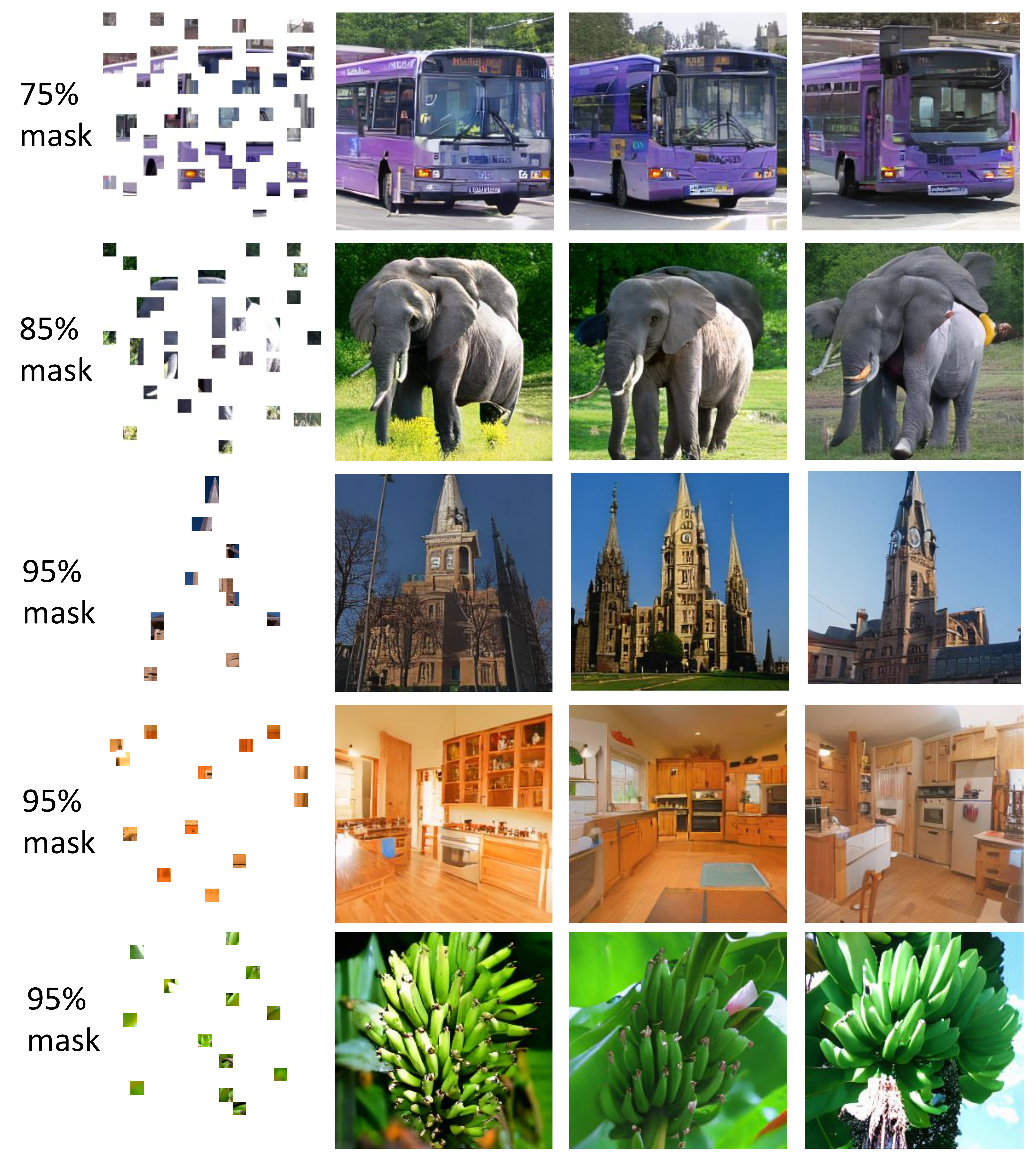}
	\caption{More samples of the Image Completion (I2I) task generated by NÜWA.}
	\label{fig:supp_s2i_2}
\end{figure*}

\begin{figure*}[h]
	\centering
	\includegraphics[width=\textwidth]{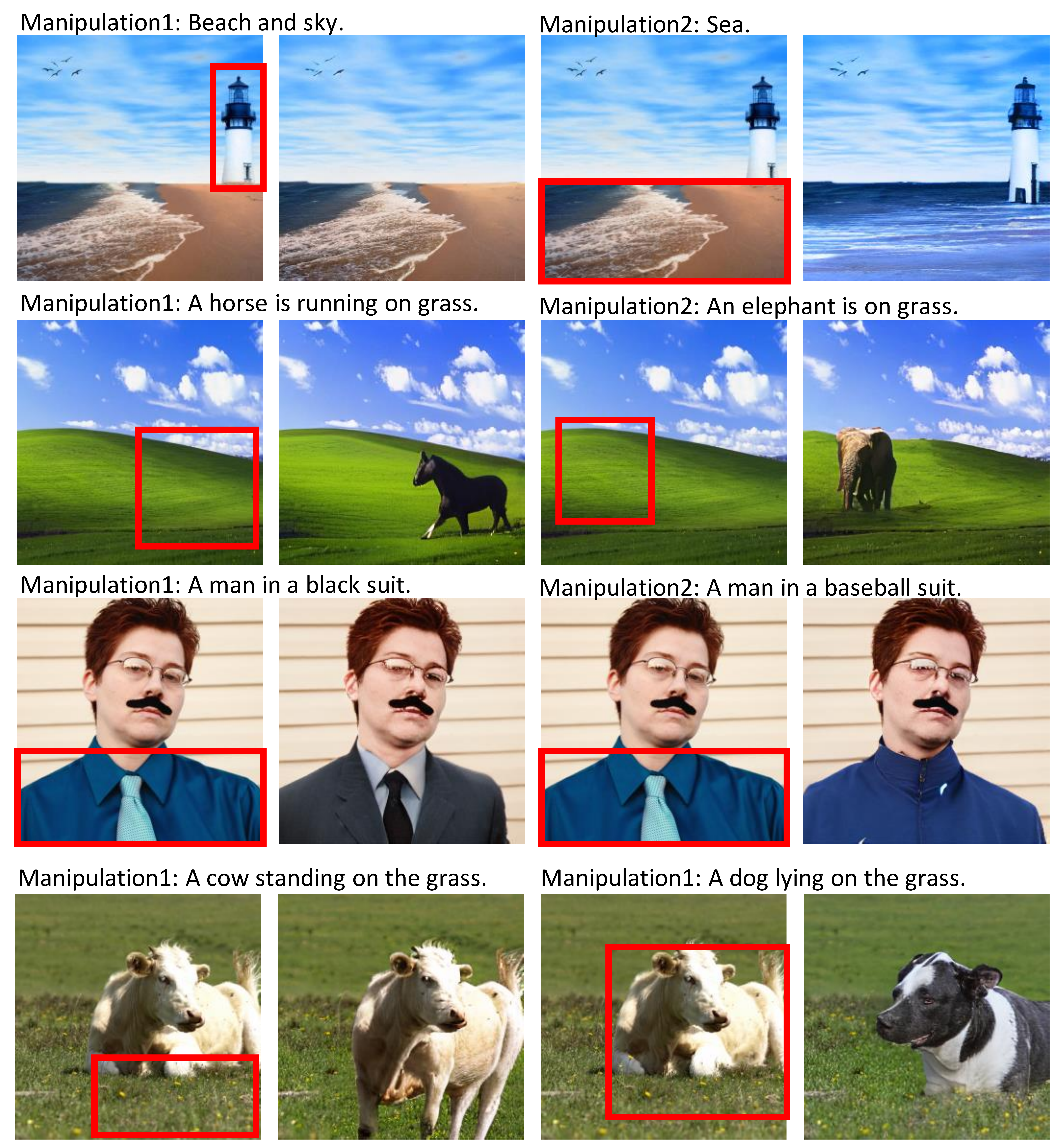}
	\caption{More samples of the Text-Guided Image Manipulation(TI2I) task generated by NÜWA.}
	\label{fig:supp_ti2i_1}
\end{figure*}

\begin{figure*}[h]
	\centering
	\includegraphics[width=\textwidth]{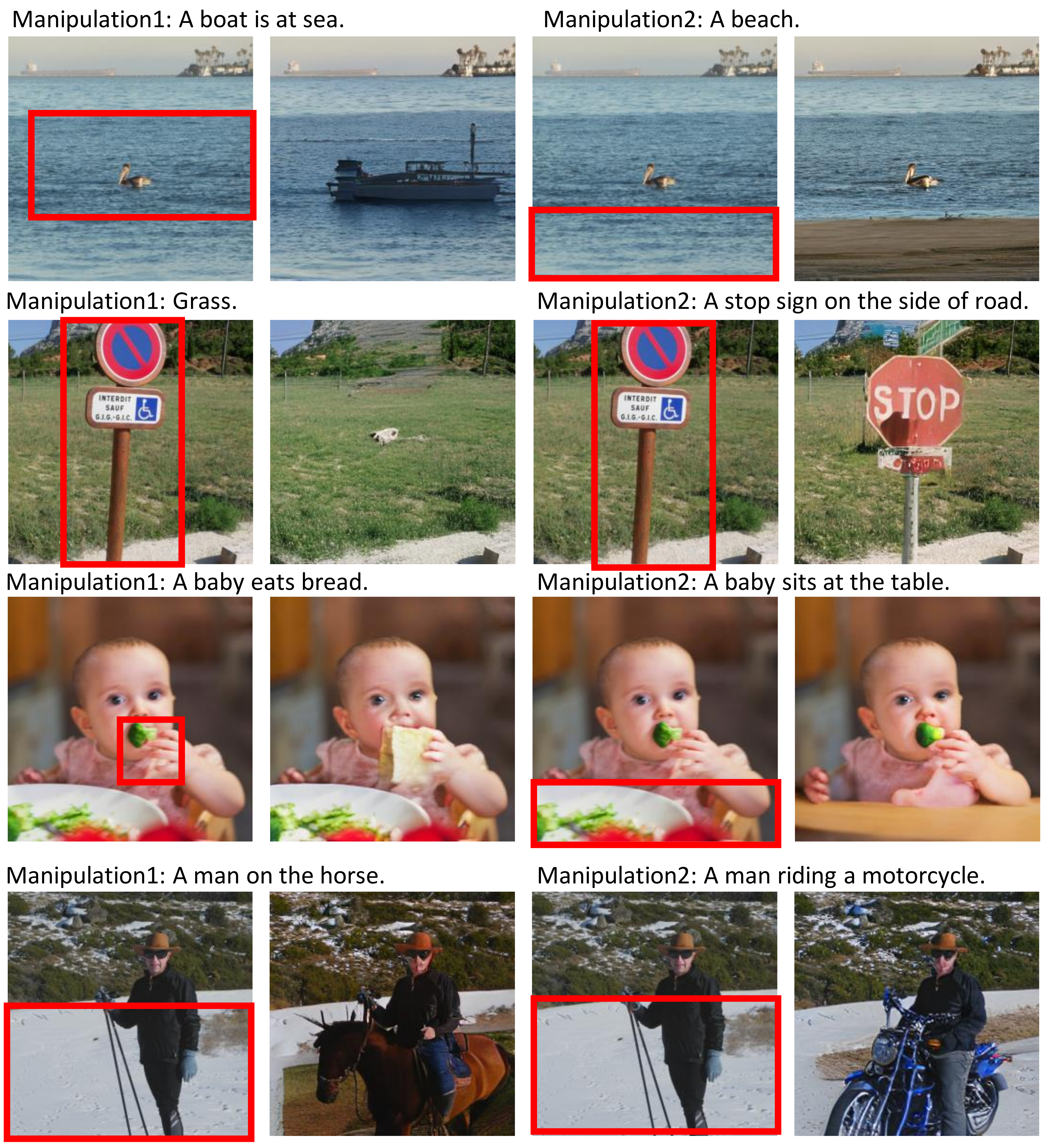}
	\caption{More samples of the Text-Guided Image Manipulation(TI2I) task generated by NÜWA.}
	\label{fig:supp_ti2i_2}
\end{figure*}

\begin{figure*}[h]
	\centering
	\includegraphics[width=\textwidth]{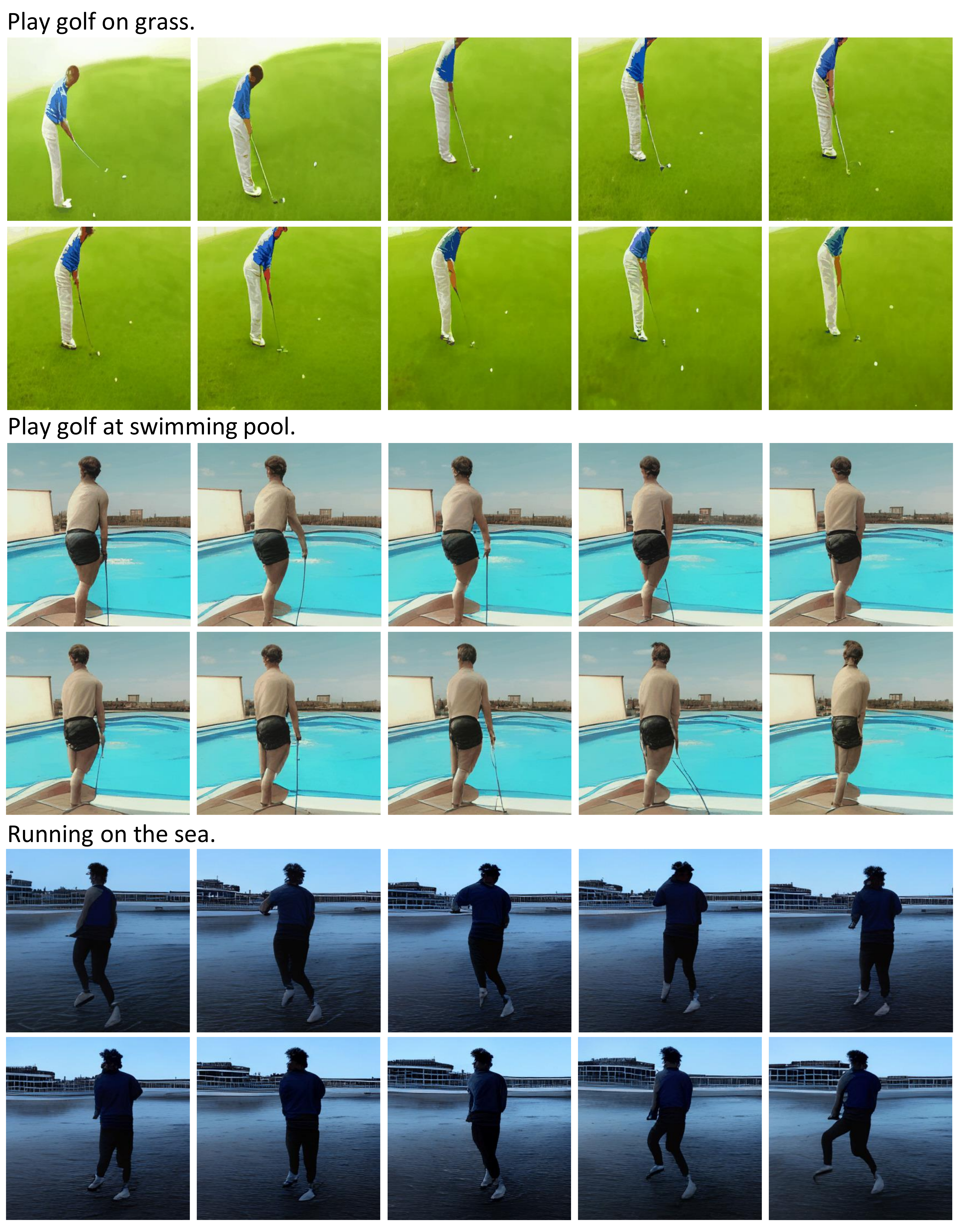}
	\caption{More samples of the Text-to-Video (T2V) task generated by NÜWA.}
	\label{fig:supp_v2v_1}
\end{figure*}

\begin{figure*}[h]
	\centering
	\includegraphics[width=\textwidth]{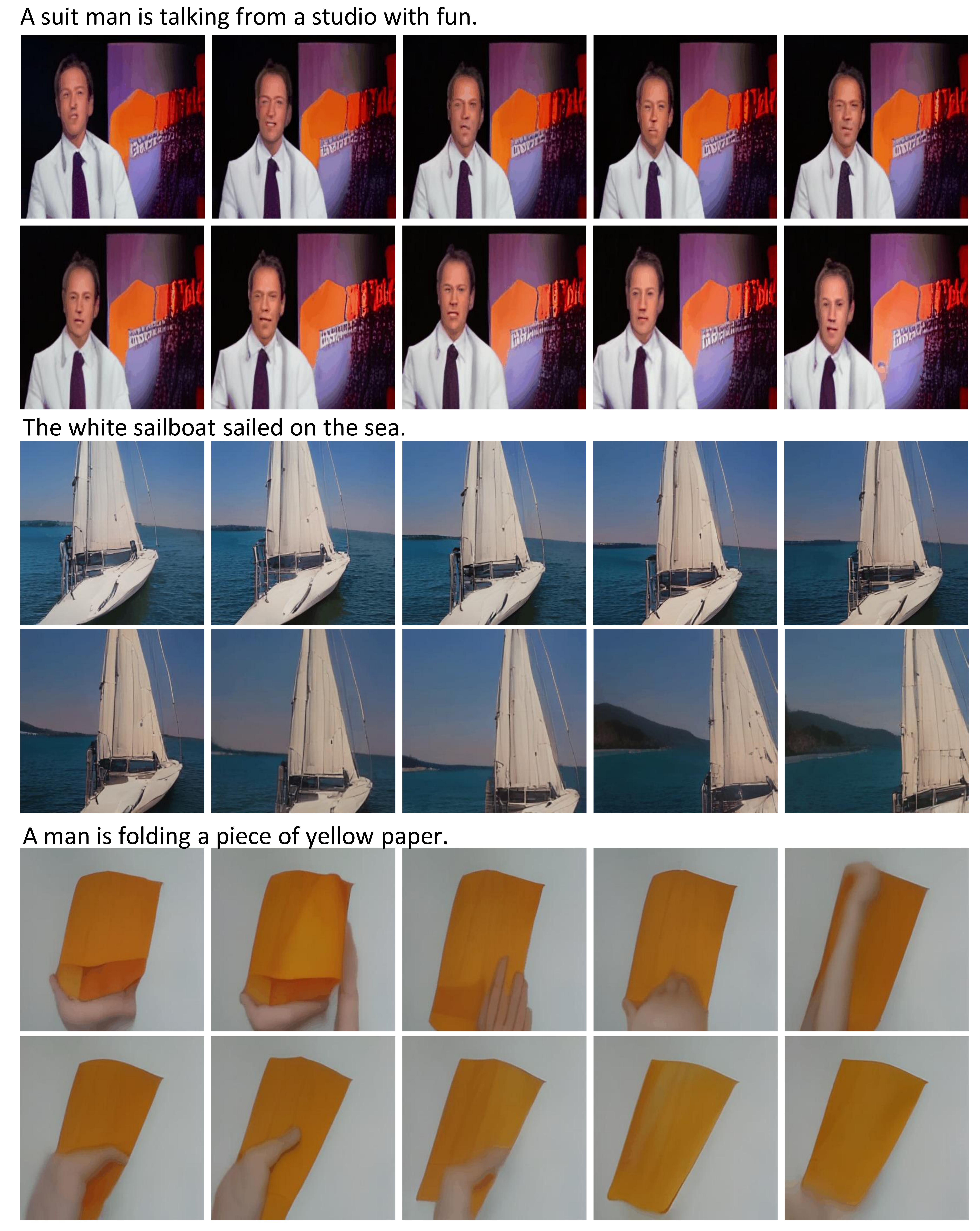}
	\caption{More samples of the Text-to-Video (T2V) task generated by NÜWA.}
	\label{fig:supp_v2v_2}
\end{figure*}
\begin{figure*}[h]
	\centering
	\includegraphics[width=0.9\textwidth]{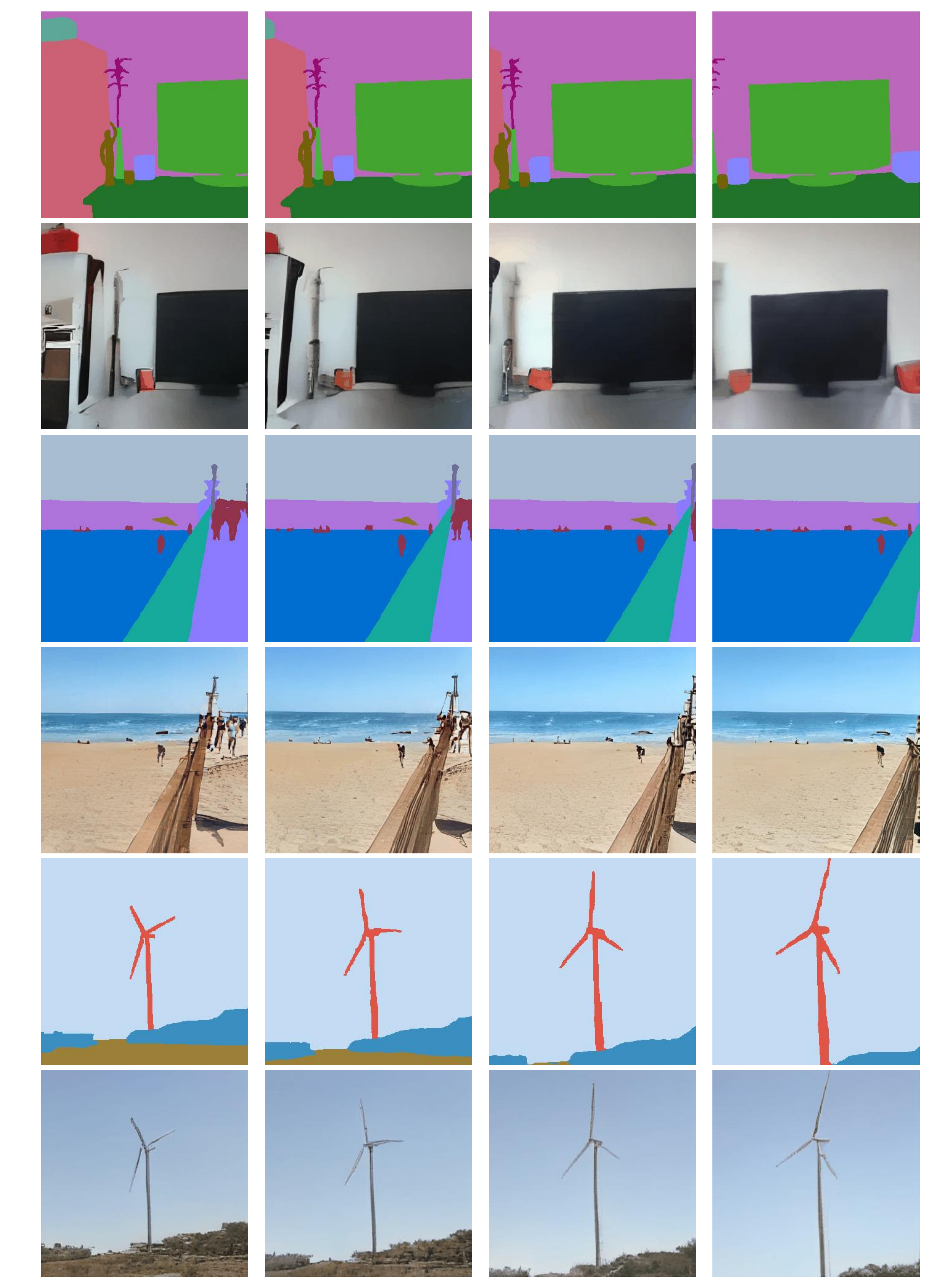}
	\caption{More samples of Sketch-to-Video (S2V) task generated by NÜWA.}
	\label{fig:supp_v2v_1}
\end{figure*}
\begin{figure*}[h]
	\centering
	\includegraphics[width=0.9\textwidth]{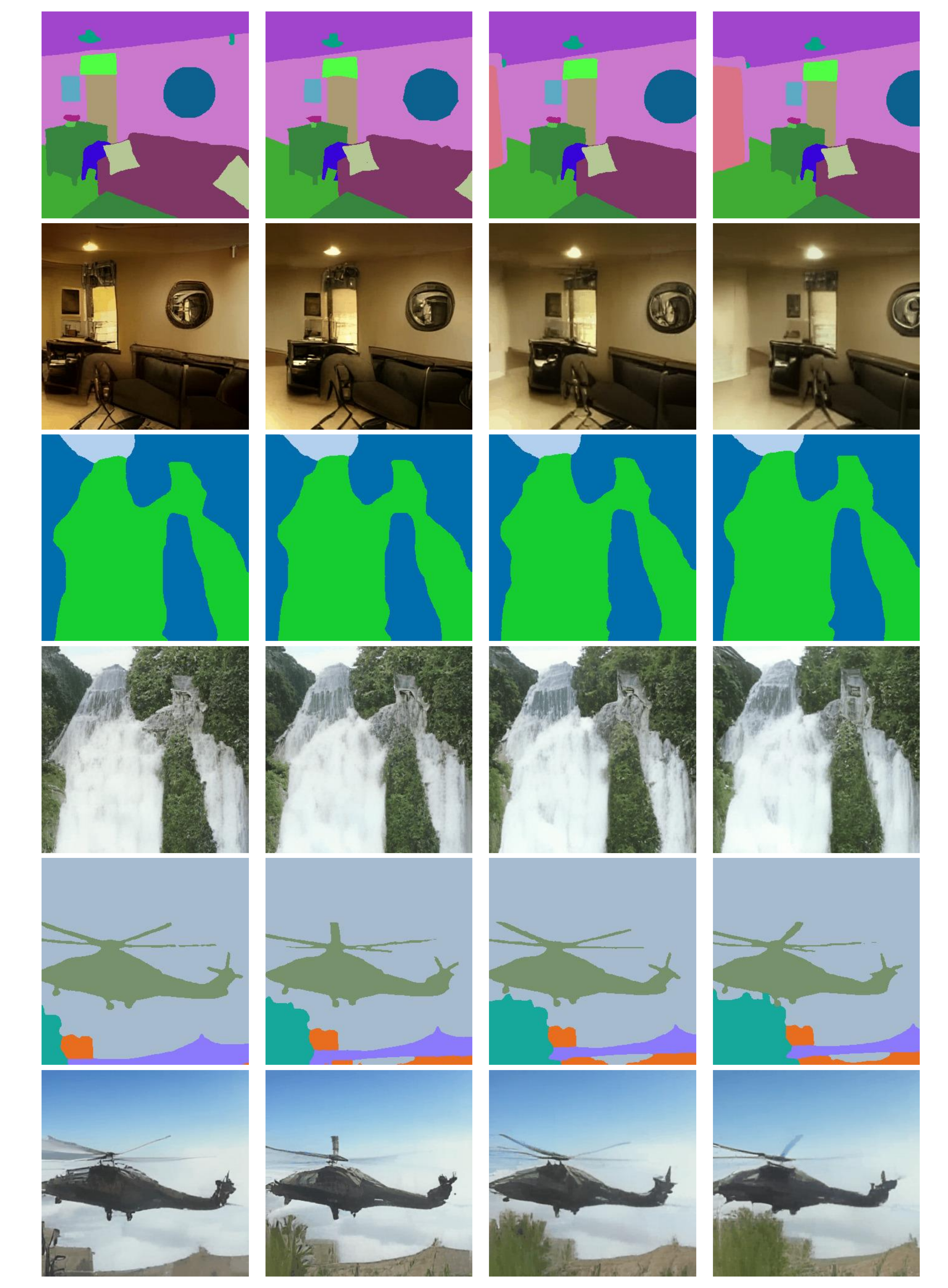}
	\caption{More samples of Sketch-to-Video (S2V) task generated by NÜWA.}
	\label{fig:supp_v2v_1}
\end{figure*}
\begin{figure*}[h]
	\centering
	\includegraphics[width=\textwidth]{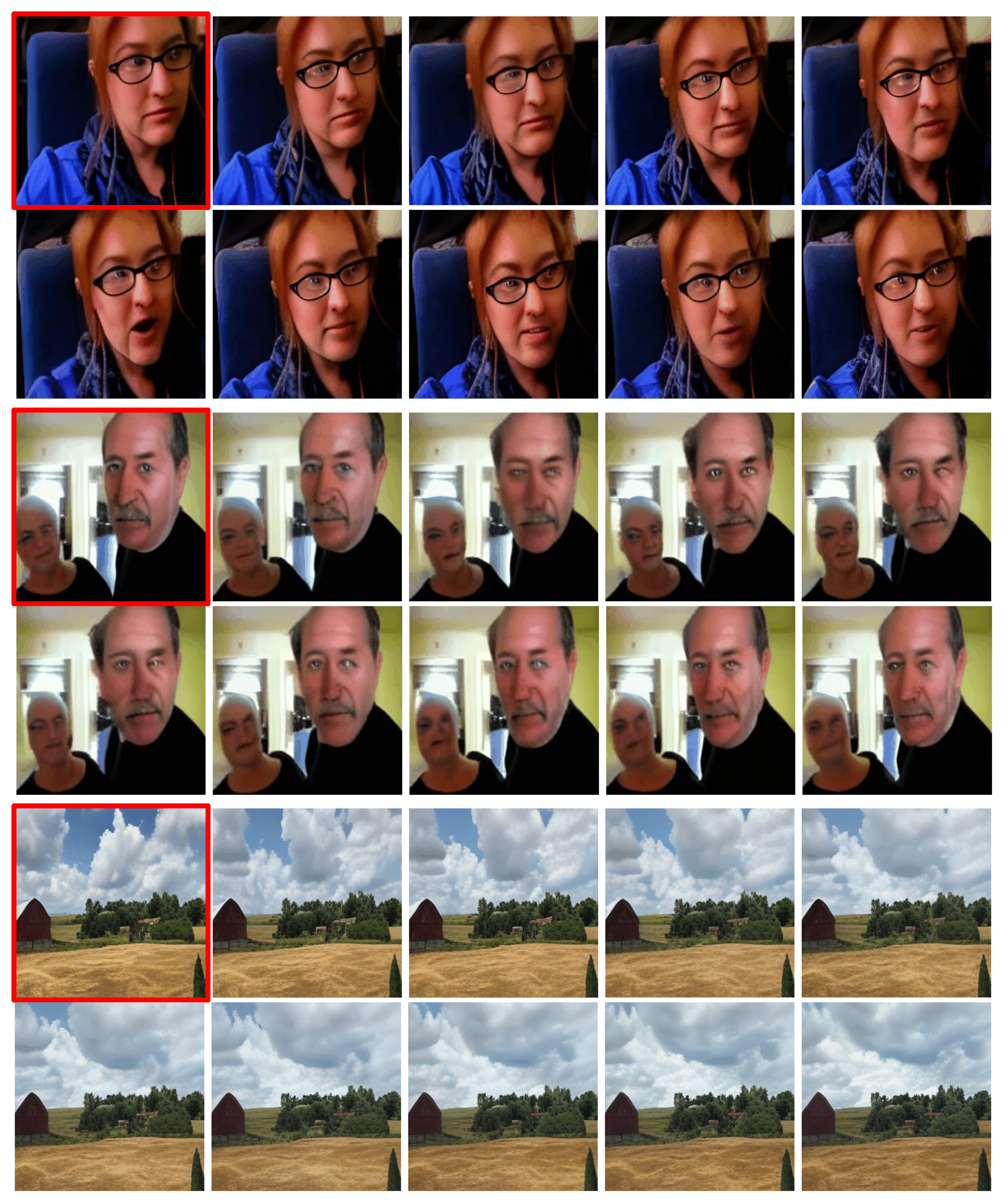}
	\caption{More samples of the Video Prediction (V2V) task generated by NÜWA. Only one frame (see red box) is used as condition.}
	\label{fig:supp_v2v_1}
\end{figure*}

\begin{figure*}[h]
	\centering
	\includegraphics[width=\textwidth]{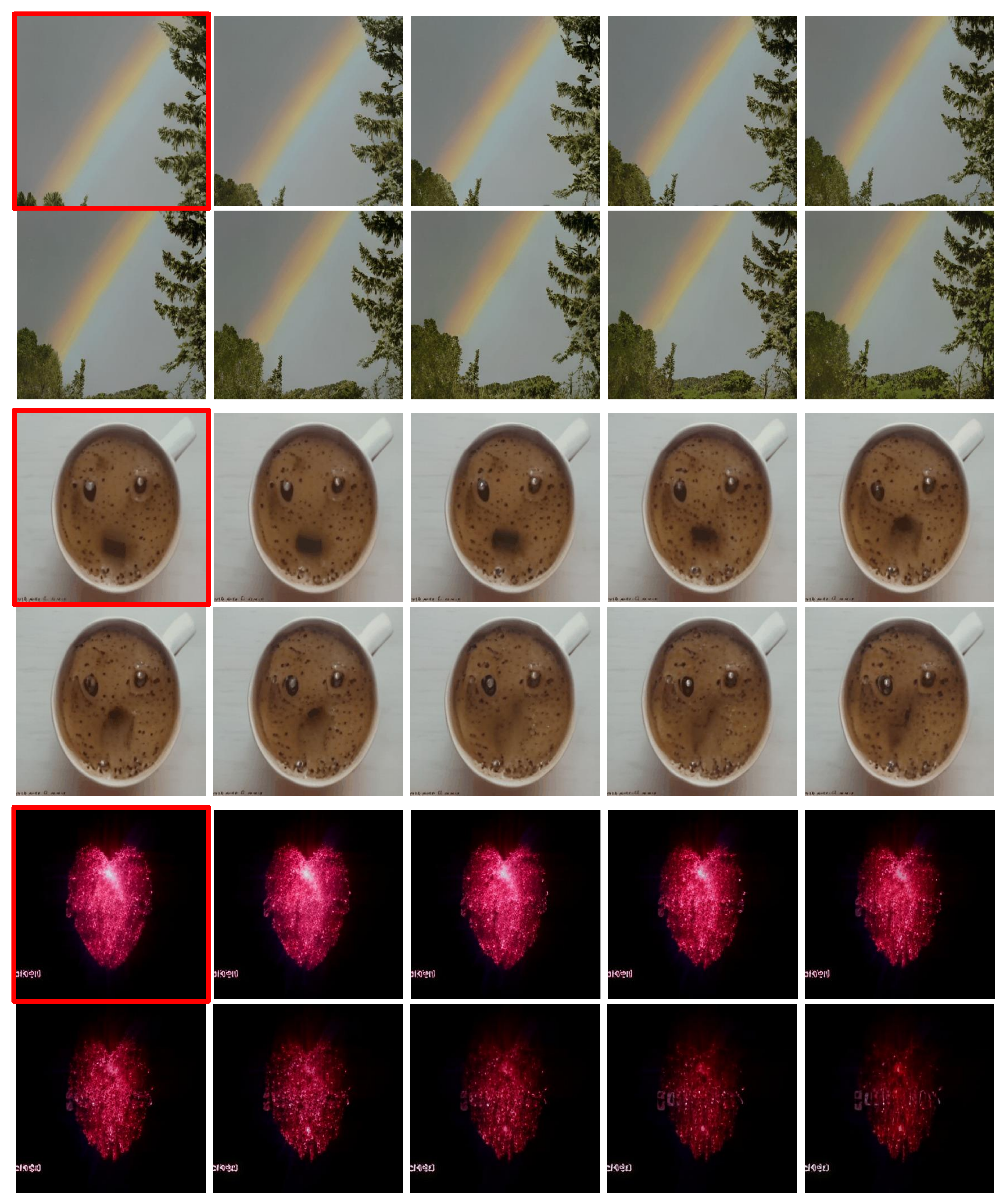}
	\caption{More samples of the Video Prediction (V2V) task generated by NÜWA.Only one frame (see red box) is used as condition.}
	\label{fig:supp_v2v_2}
\end{figure*}

\begin{figure*}[h]
	\centering
	\includegraphics[width=\textwidth]{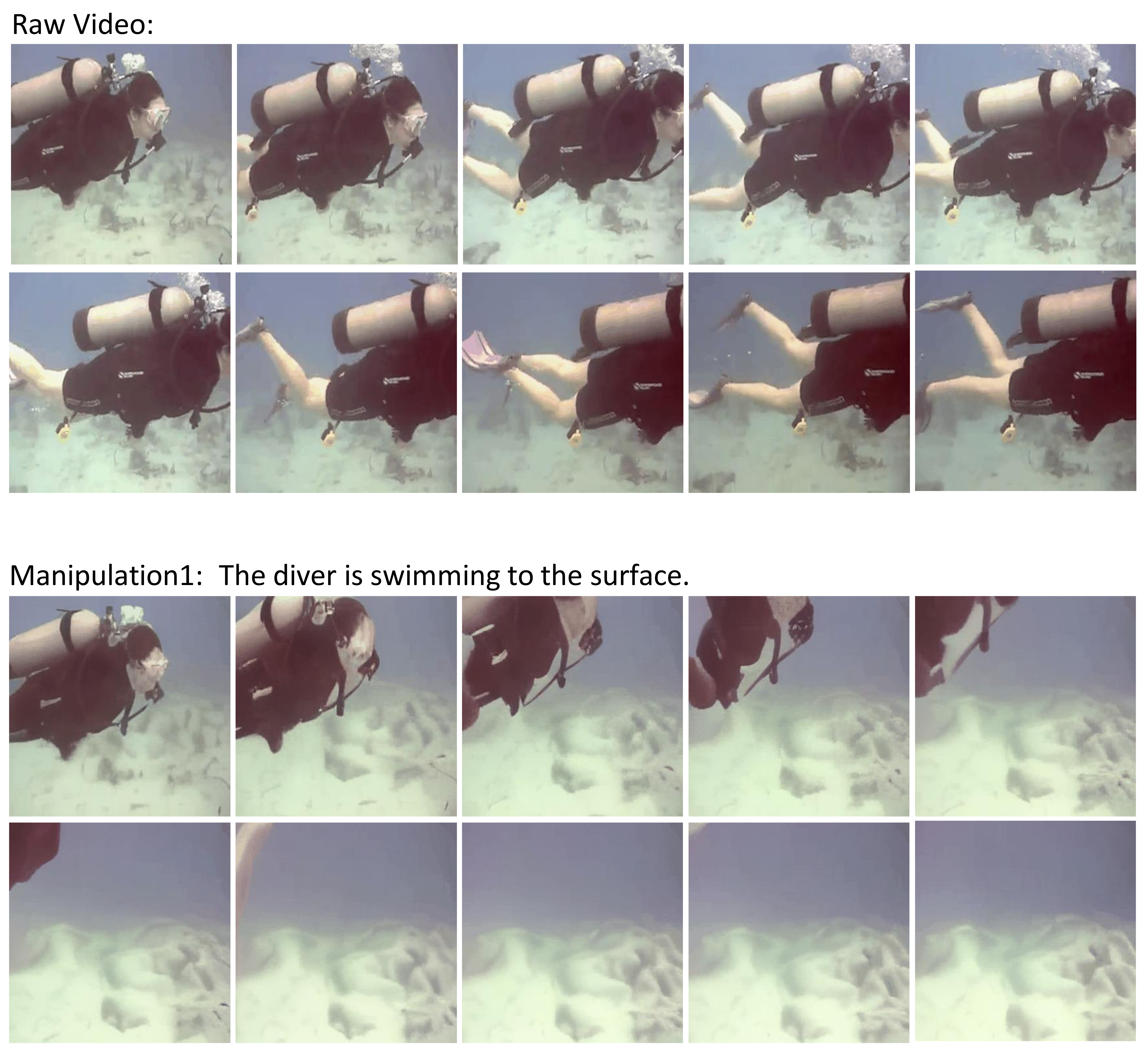}
	\caption{More samples of Text-Guided Video Manipulation (TV2V) task generated by NÜWA.}
	\label{fig:supp_v2v_2}
\end{figure*}

\begin{figure*}[h]
	\centering
	\includegraphics[width=\textwidth]{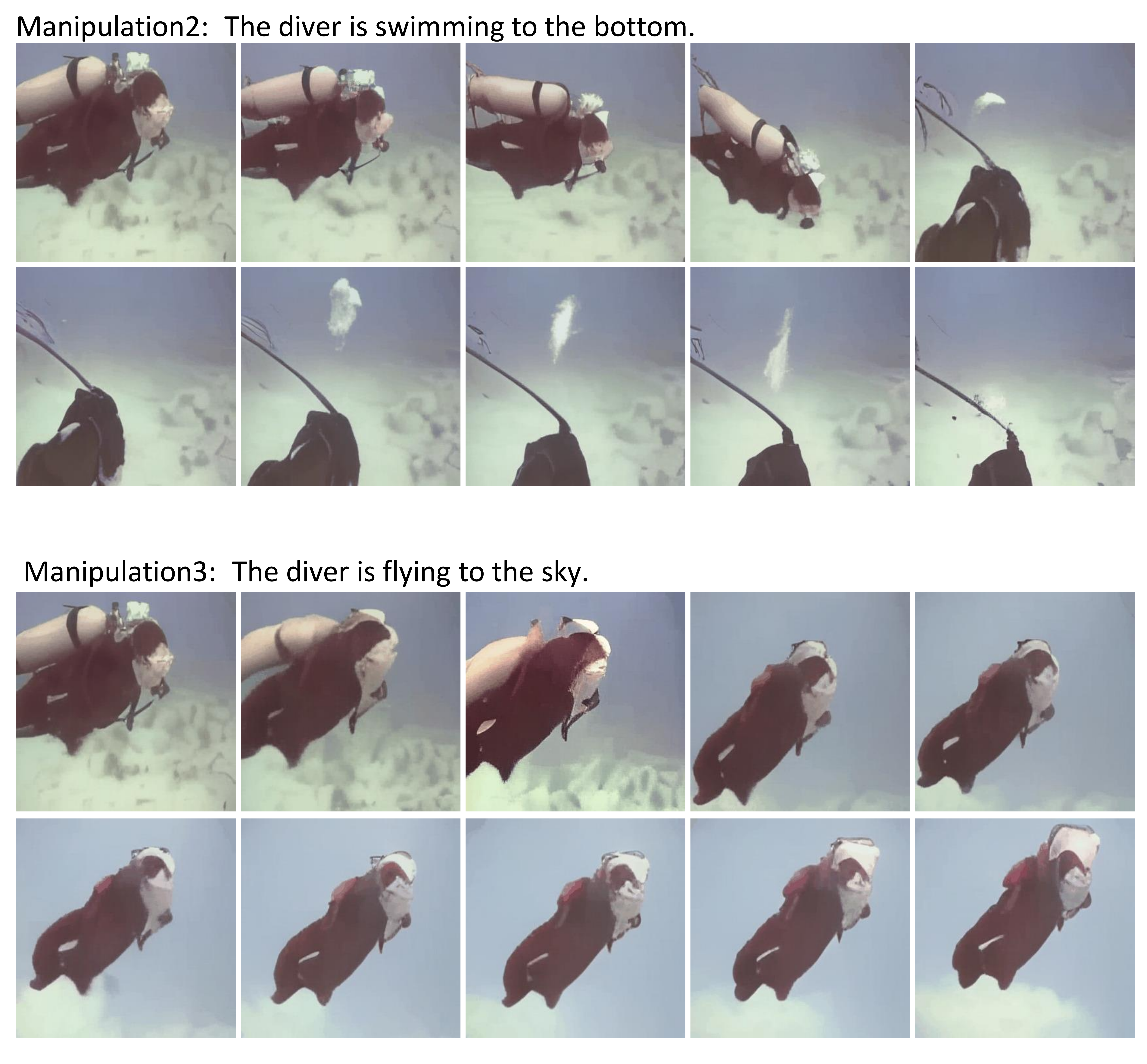}
	\caption{More samples of Text-Guided Video Manipulation (TV2V) task generated by NÜWA.}
	\label{fig:supp_v2v_2}
\end{figure*}

\end{document}